\documentclass[10pt,twocolumn,letterpaper]{article}

\usepackage{amsmath,amsfonts,bm}

\def\eqref#1{equation~\ref{#1}}

\def\1{\bm{1}}

\def\rvepsilon{{\mathbf{\epsilon}}}

\def\vs{{\bm{s}}}

\def\vx{{\bm{x}}}

\DeclareMathAlphabet{\mathsfit}{\encodingdefault}{\sfdefault}{m}{sl}
\SetMathAlphabet{\mathsfit}{bold}{\encodingdefault}{\sfdefault}{bx}{n}

\usepackage{iccv}
\usepackage{times}
\usepackage{epsfig}
\usepackage{graphicx}
\usepackage{amsmath}
\usepackage{amssymb}
\usepackage{booktabs}
\usepackage[table]{xcolor}
\usepackage[super]{nth}
\usepackage{enumitem}
\usepackage[english]{babel}
\usepackage{blindtext}
\usepackage[pagebackref,breaklinks=true,colorlinks,citecolor=citecolor2,linkcolor=linkcolor,bookmarks=false]{hyperref}

\usepackage[capitalize]{cleveref}
\crefname{section}{Sec.}{Secs.}
\Crefname{section}{Section}{Sections}
\Crefname{table}{Table}{Tables}
\crefname{table}{Tab.}{Tabs.}

\newlength\savewidth\newcommand\shline{\noalign{\global\savewidth\arrayrulewidth
\global\arrayrulewidth 1pt}\hline\noalign{\global\arrayrulewidth\savewidth}}
\newcommand{\tablestyle}[2]{\setlength{\tabcolsep}{#1}\renewcommand{\arraystretch}{#2}\centering\small}
\makeatletter\renewcommand\paragraph{\@startsection{paragraph}{4}{\z@}
{.4em \@plus1ex \@minus.2ex}{-.5em}{\normalfont\normalsize\bfseries}}\makeatother

\newcolumntype{x}[1]{>{\centering\arraybackslash}p{#1pt}}
\newcolumntype{y}[1]{>{\raggedright\arraybackslash}p{#1pt}}
\newcolumntype{z}[1]{>{\raggedleft\arraybackslash}p{#1pt}}

\def\ourmodel{{DiffMAE}\xspace}
\def\x{$\times$}
\definecolor{citecolor}{RGB}{34,139,34}
\definecolor{citecolor2}{HTML}{0071bc}
\definecolor{lightred}{RGB}{241,140,142}
\definecolor{defaultcolor}{gray}{0.9}
\definecolor{demphcolor}{gray}{.7}
\definecolor{linkcolor}{HTML}{ED1C24}

\newcommand{\demph}[1]{\textcolor{demphcolor}{#1}}
\newcommand{\app}{\raise.17ex\hbox{$\scriptstyle\sim$}}

\begin{document}

\title{Diffusion Models as Masked Autoencoders}

\author{Chen Wei\textsuperscript{1 2}\quad Karttikeya Mangalam\textsuperscript{1}\quad Po-Yao Huang\textsuperscript{1}\quad Yanghao Li\textsuperscript{1}  \\  Haoqi Fan\textsuperscript{1}\quad Hu Xu\textsuperscript{1}\quad Huiyu Wang\textsuperscript{1}\quad Cihang Xie\textsuperscript{3}\quad Alan Yuille\textsuperscript{2}\quad Christoph Feichtenhofer\textsuperscript{1} \vspace{.5em} \\
\textsuperscript{1}FAIR, Meta AI \qquad \textsuperscript{2}Johns Hopkins University 
 \qquad \textsuperscript{3}UC Santa Cruz
}
\maketitle

\begin{abstract}
There has been a longstanding belief that generation can facilitate a true understanding of visual data. In line with this, we revisit generatively pre-training visual representations in light of recent interest in denoising diffusion models. While directly pre-training with diffusion models does not produce strong representations, we condition diffusion models on masked input and formulate diffusion models as masked autoencoders (\ourmodel). Our approach is capable of (\textit{i}) serving as a strong initialization for downstream recognition tasks, (\textit{ii}) conducting high-quality image inpainting, and (\textit{iii}) being effortlessly extended to video where it produces state-of-the-art classification accuracy. We further perform a comprehensive study on the pros and cons of design choices and build connections between diffusion models and masked autoencoders. \href{https://weichen582.github.io/diffmae.html}{Project page.}

\vspace{-8pt}
\end{abstract}

\section{Introduction}
\label{sec:intro}

\begin{flushleft}
  \vspace{-5pt}
  ``\textit{What I cannot create, I do not understand.}'' \vspace{0.05in}
  \\\raggedleft{------ Richard P. Feynman, 1988} 
  \vspace{-5pt}
\end{flushleft}
For many years, there has been a desire to achieve a deeper understanding of visual data through the process of generation.
Early approaches, such as deep belief networks~\cite{dbn} and denoising autoencoders~\cite{denoisingae}, employed generative pre-training to initialize deep networks for downstream recognition tasks. As generative models can \textit{create} new samples by approximating the data distribution, it stands to reason that such modeling should simultaneously arrive at a semantic \textit{understanding} of the raw visual data, as required for recognition tasks, following Feynman.

In line with this philosophy, generative language models, \eg, Generative Pre-trained Transformers or GPTs~\cite{gpt-3}, obtain a broad understanding of language and an immense knowledge base, excelling as both a few-shot learner and a pre-trained base model. Nevertheless, recent explorations in vision generative pre-training fall out of favor. For example, GAN-based BiGAN~\cite{bigan,bigbigan} and auto-regressive iGPT~\cite{igpt} noticeably fall short of their concurrent contrastive algorithms despite of using 10$\times$ more parameters. The challenge stems, in part, from the divergent focus: While recognition models primarily focus on the high-level low-frequency structure of images, generation models must also allocate capacity for low-level high-frequency details~\cite{dall-e}. Given this discrepancy, it remains an open question whether and how generative pre-training can effectively compete with other self-supervised algorithms on downstream recognition tasks despite its intuitive appeal.

\begin{figure}[!t]
\centering
\includegraphics[width=\linewidth]{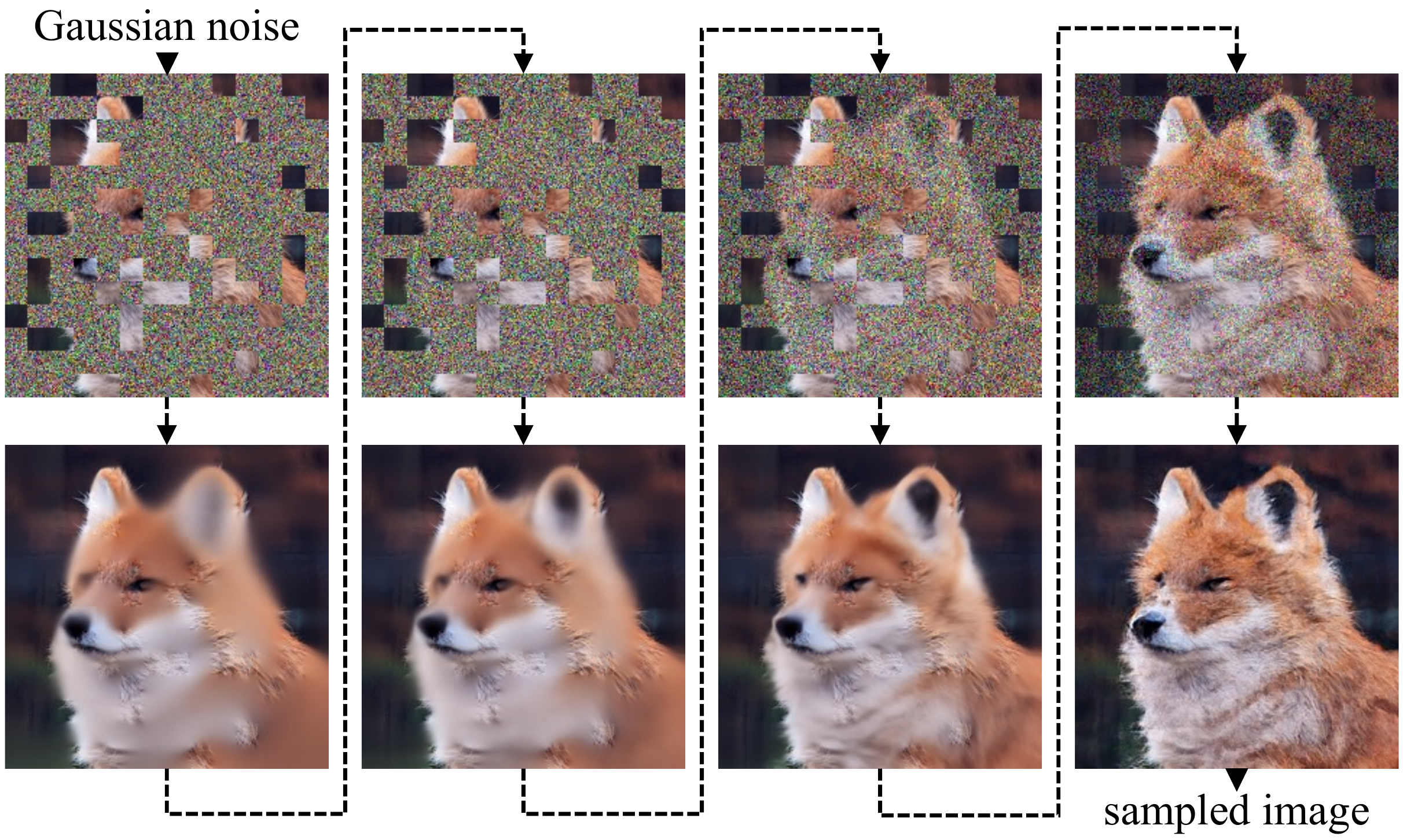}
\vspace{-15pt}
\caption{\textbf{Inference process of \ourmodel}, which iteratively unfolds from random Gaussian noise to the sampled output. During training, the model learns to denoise the input at different noise levels (from \textit{top} row to the \textit{bottom}) and simultaneously performs self-supervised pre-training for downstream recognition.}
\label{fig:teaser}
\vspace{-10pt}
\end{figure}

In recent years, the field of image generation has been dominated by denoising diffusion models~\cite{sohl2015deep, ddpm}. These models employ a straightforward process of iteratively refining noisy samples. As a result, the generated images are of impressively high quality, and even better, can generate an extensive range of diverse samples~\cite{glide, dalle-2, imagen}. In light of this progress, we revisit the potential of generative pre-training in the context of diffusion models.

To begin with, we directly fine-tune a pre-trained diffusion model~\cite{adm} on ImageNet classification~\cite{imagenet}. Despite its strong performance for unconditional image generation, the pre-trained diffusion model only yields a marginal improvement in classification when compared to training the same architecture from scratch (\cref{sec:finetuning-diffuison}), and is outperformed by concurrent self-supervised pre-training algorithms such as Masked Autoencoders (MAE)~\cite{mae}.

MAE demonstrates strong recognition performance by learning to regress pixels of masked patches given the other visible patches. Inspired by MAE, we incorporate masking into diffusion models and cast Diffusion Models as Masked Autoencoders (\ourmodel). We formulate the masked prediction task as a \textit{conditional} generative objective, \ie, to approximate the pixel distribution of the masked region conditioned on the visible region. We learn models with our diffusion approach, within the MAE framework, introducing no extra training cost. During pre-training, our model is trained to denoise the input at different noise levels and learns a strong representation for \textit{recognition and generation}. We evaluate the pre-trained model by fine-tuning on downstream recognition tasks, as well as image inpainting for which the model generates samples by iteratively unfolding from random Gaussian noise, illustrated in \cref{fig:teaser}.

The diffusion nature of \ourmodel allows it to generate intricate visual details, \eg, of objects (\cref{fig:main-viz}). In contrast, MAE is known to produce blurry reconstructions that lack high-frequency components. Further, \ourmodel maintains strong performance on image and video recognition tasks.

We make the following observations in this work:

(\textit{i}) \ourmodel is a strong pre-training approach for fine-tuning on downstream \textit{recognition} tasks, obtaining comparable performance to leading 
self-supervised learning algorithms that focus solely on recognition. When combined with CLIP~\cite{clip} features, our \ourmodel is even able to outperform recent work that combines MAE and CLIP.

(\textit{ii}) \ourmodel is able to \textit{generate} high quality images conditioning on masked input. Specifically, \ourmodel generations outperform leading inpainting methods quantitatively and also appear more semantically meaningful.  

(\textit{iii}) \ourmodel can be extended to the \textit{video} domain effortlessly, providing high-quality inpainting and state-of-the-art recognition accuracy, outperforming recent works. 

(\textit{iv}) We reveal a \textit{connection} between MAE and diffusion models, as MAE effectively performs the first inference step of diffusion. In other words, we believe the success of MAE aligns with the philosophy of generation for recognition.  

We further perform a comprehensive empirical study to elucidate the pros and cons of the design choices on downstream recogntion and inpainting generation tasks.

\section{Related Work}
\label{sec:related}

\paragraph{Self-supervised learning} aims to learn from unlabeled visual data by a pre-text task that is constructed by image/patch operations (\eg, \cite{context-prediction,jigsaw,jigsaw2,colorization,rotation,deepclustering}) and spatial-temporal operations (\eg, \cite{goroshin2015unsupervised,misra2016shuffle,fernando2017self,pathak2017learning,wang2017transitive}). Contrastive methods~\cite{dosovitskiy2015discriminative} capitalize on augmentation invariance of images and videos~\cite{instdist,moco,simclr,swav,byol,dino,simsiam,cvrl,videomoco}.

For vision, different masked prediction targets have been proposed. MAE~\cite{mae} predicts pixel colors with an efficient asymmetric architecture. BEiT~\cite{beit, beitv2} and iBOT~\cite{ibot} predict dVAE~\cite{vqvae,dall-e} or learnable tokens. \mbox{MaskFeat}~\cite{maskfeat} predicts HOG features. data2vec~\cite{data2vec} learns from a momentum teacher. The community is continuously exploring this direction~\cite{peco,simmim,hu2022exploring,zhang2022hivit,dbot,videomae}.

\paragraph{Generative learning for recognition} has a long-standing appeal for its intuitiveness. Pioneers study the representation learned by GANs~\cite{gan,dcgan,bigan} and VAEs~\cite{kingma2013auto,kingma2014semi}. BigBiGAN~\cite{bigbigan} demonstrates a model that learns competitive recognition representation and generates high-fidelity images with GAN. This work is followed by iGPT~\cite{igpt}, which generates images autoregressively and was state-of-the-art on linear probing protocols. Though this stream stagnates afterward, we demonstrate its resurgence by exploring diffusion models in this work.

\paragraph{Denoising diffusion models} have ushered in a new era of diverse, high-resolution conditional image generation~\cite{sohl2015deep, adm, ddpm}. Utilizing a forward Gaussian diffusion process and a backward generation process, denoising diffusion models iteratively refine the generated image starting from Gaussian noise. This process has proven extremely powerful for rich text-conditioned generation of both images~\cite{dalle-2, glide, imagen, stable} and videos~\cite{phenaki, ho2022video, imagenvideo, make-a-video}. 

\paragraph{Masked autoencoders} were pioneered by stacked autoencoders~\cite{denoisingae} and inpainting tasks~\cite{inpainting} using ConvNets. Since the introduction of ViT~\cite{vit}, masked prediction has re-attracted attention, partially inspired by the success of BERT~\cite{bert} in NLP. BERT performs masked language modeling, scales well, and generalizes to different end tasks.

\section{Fine-tuning Generative Models}
\label{sec:finetuning-diffuison}

\begin{table}[!h]
\vspace{-10pt}
\centering
\tablestyle{2pt}{1.02}
\begin{tabular}{x{50}|x{50}x{35}x{35}x{40}}
pre-train         & architecture    & params. & scratch & pre-trained \\
\shline
\demph{MAE~\cite{mae}}   & \demph{ViT-L}   & \demph{304M}  & \demph{82.6}\textcolor{white}{$^\dagger$} & \demph{85.9}   \\
\hline
iGPT~\cite{igpt}         & iGPT-L          & 1362M         & 53.2\textcolor{white}{$^\dagger$}             & 72.6  \\
ADM~\cite{adm}           & U-Net enc.      & 211M          & 82.0$^\dagger$   & 83.3  \\
DDPM~\cite{ddpm}         & ViT-L           & 304M          & 82.6\textcolor{white}{$^\dagger$} & 83.4 \\
\textbf{\ourmodel}       & ViT-L           & 304M          & 82.6\textcolor{white}{$^\dagger$} & \textbf{85.8}   \\
\end{tabular}
\vspace{2pt}
\caption{\textbf{Fine-tuning generative models on ImageNet}, a system-level comparison. $^\dagger$trained from scratch by us. Non-generative method is included for reference and \demph{de-emphasized}.}
\label{tab:gen-models}
\vspace{-10pt}
\end{table}

\begin{figure*}[!ht]
\vspace{-5pt}
\centering
\tablestyle{0.3pt}{0.2}
\begin{tabular}{cc}
\includegraphics[width=0.5\textwidth]{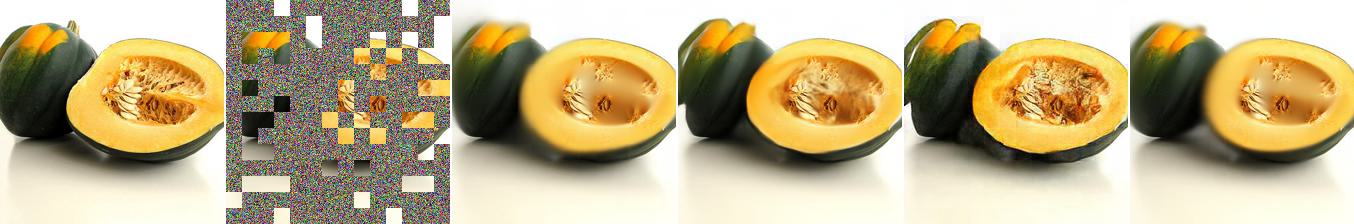} &
\includegraphics[width=0.5\textwidth]{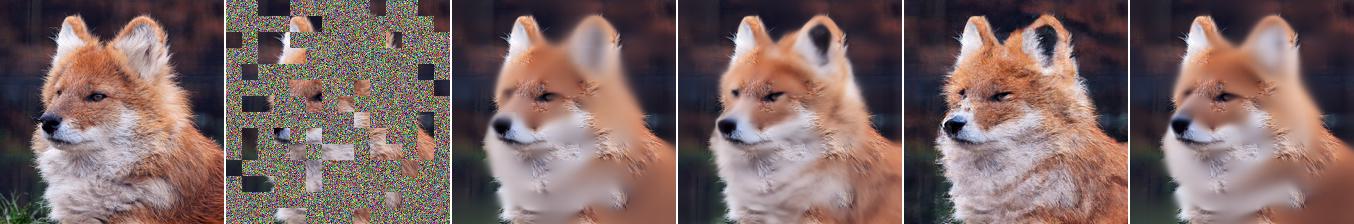} \\
\includegraphics[width=0.5\textwidth]{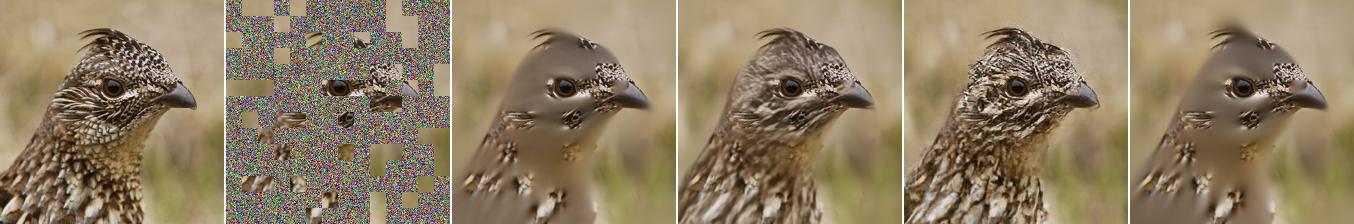} &
\includegraphics[width=0.5\textwidth]{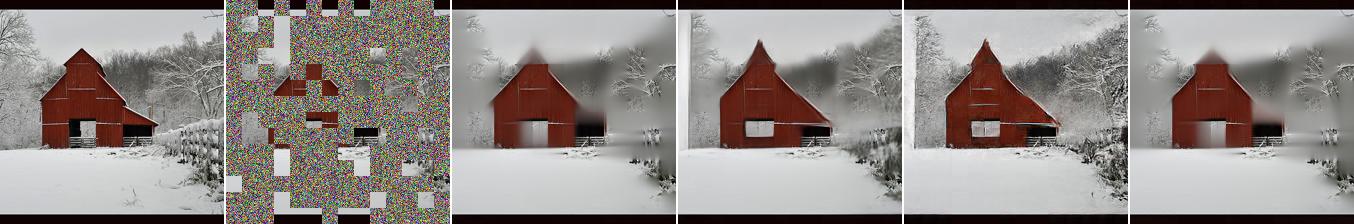} \\
\includegraphics[width=0.5\textwidth]{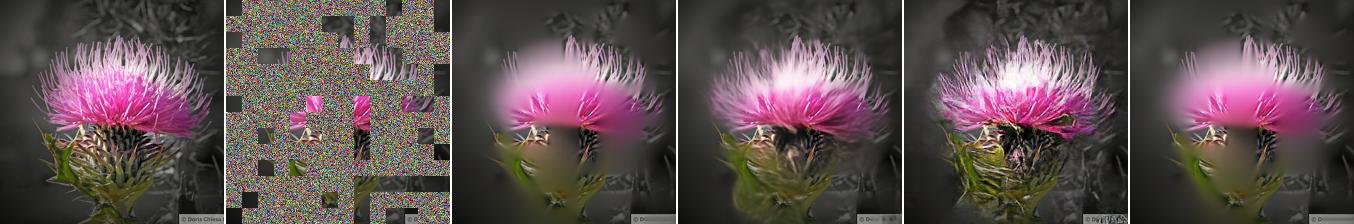} &
\includegraphics[width=0.5\textwidth]{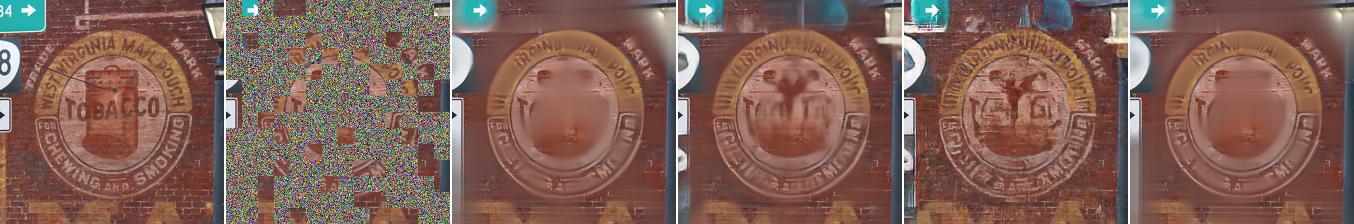} \\
\includegraphics[width=0.5\textwidth]{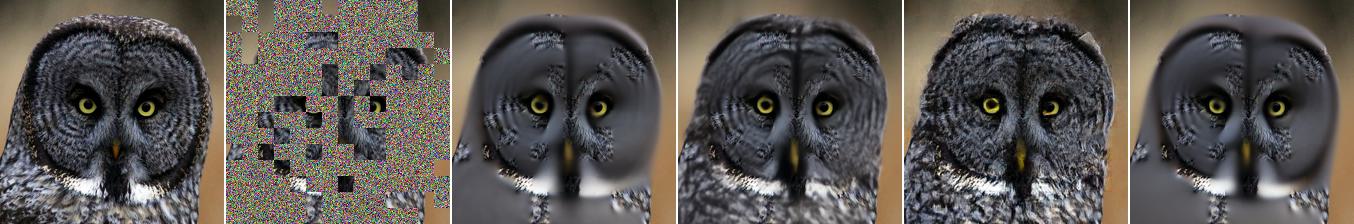} &
\includegraphics[width=0.5\textwidth]{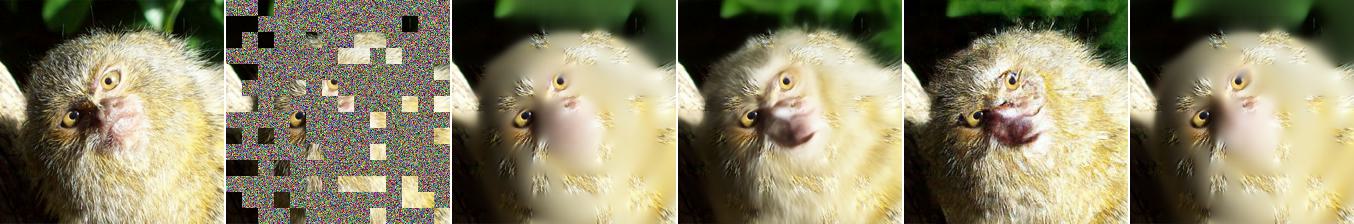} \\
\includegraphics[width=0.5\textwidth]{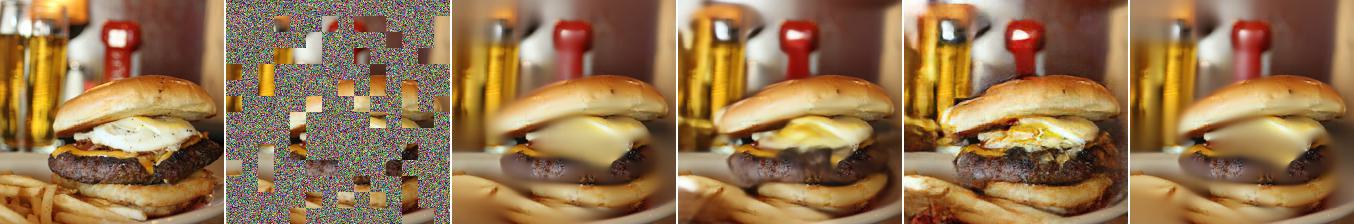} &
\includegraphics[width=0.5\textwidth]{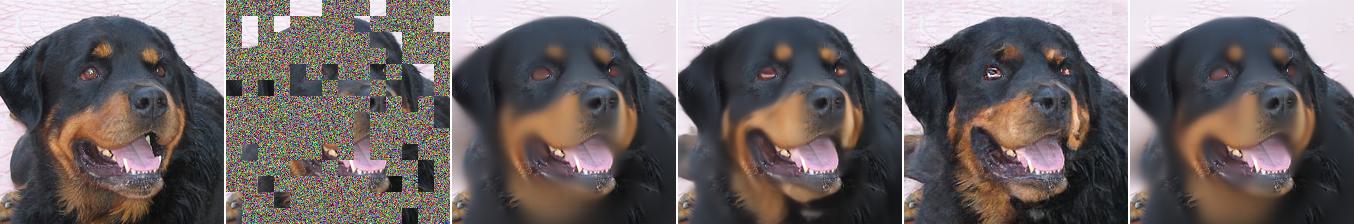} \\
\end{tabular}

\tablestyle{0pt}{1.0}
\begin{tabular}{x{41}x{41}x{41}x{41}x{41}x{42}x{41}x{41}x{41}x{41}x{41}x{41}}
\footnotesize{ground-truth} & \footnotesize{input} & \footnotesize{$t=1000$} & \footnotesize{$t=500$} & \footnotesize{\textbf{$t=0$}} & \footnotesize{MAE} & \footnotesize{ground-truth} & \footnotesize{input} & \footnotesize{$t=1000$} & \footnotesize{$t=500$} & \footnotesize{\textbf{$t=0$}} & \footnotesize{MAE}
\end{tabular}
\caption{\textbf{Qualitative comparison of \ourmodel and MAE}. $t$\,$=$\,$1000$ and $t$\,$=$\,$500$ are \ourmodel's predictions at intermediate timesteps from $t$\,$=$\,$1000\ldots0$. $t$\,$=$\,$0$ marks the final generation. MAE's outputs are obtained from its official visualization model. The predictions at the masked area and the ground-truth at the visible area are overlaid. Both models are ViT-L pre-trained for 1600 epochs.}
\label{fig:main-viz}
\vspace{-10pt}
\end{figure*}

We start by comparing different generative pre-training methods for downstream ImageNet classification in \cref{tab:gen-models}.

iGPT~\cite{igpt} generates images in a GPT~\cite{gpt-3} style by auto-regressively predicting pixels. iGPT pre-training has significantly improved its low from-scratch (random initialization) accuracy from 53.2\% to 72.6\%, which is still relatively low, considering the model size (1362M) and compute cost.

To re-examine generative pre-training in light of the recent progress on diffusion models, we fine-tune a pre-trained ADM~\cite{adm}, which is a recent method for diffusion-based image generation. In particular, we take its unconditional IN-1K model, which aligns with the criteria of unsupervised pre-training and never uses the class labels. Compared to the from-scratch counterpart, initializing with the pre-trained diffusion model provides a gain of +1.3\% top-1. However, the resulting 83.4\% still lags far behind the non-generative self-supervised algorithms such as MAE (85.9\%). Considering the differences in architecture, we further train a ViT-L with diffusion DDPM~\cite{ddpm}. This model enhances the fine-tuning classification to 83.4\% compared to its from-scratch baseline of 82.6\% top-1~\cite{mae}, which aligns with our observations on fine-tuning the pre-trained ADM. Implementation details are in the Appendix.

In comparison to these approaches, a \ourmodel trained ViT-L obtains 85.8\% when fine-tuned and is described next.

\section{Diffusion Masked Autoencoder}
\label{sec:method}

We incorporate the masked prediction paradigm into diffusion models. The model approximates the pixel distribution of the masked region conditioned on the visible region. In other words, we study Diffusion Models \textit{as} a form of Masked Autoencoders (DiffMAE), introduced next. 

\subsection{Conditional Diffusion Model}
Given a training sample $\vx_0$\,$\sim$\,$p(\vx_0)$ where subscript~$_0$ denotes that the sample is original and clean, we first spatially divide $\vx_0$ into the non-overlapping masked region $\vx_0^m$ and the visible region $\vx_0^v$. We want to model the distribution of the masked region $\vx_0^m$ conditioned on the visible region $\vx_0^v$, \ie, the distribution $p(\vx_0^m | \vx_0^v)$.

In the forward process, only the masked area $\vx_0^m$ is gradually diffused, \ie, corrupted by recursively adding a small amount of Gaussian noise $T$ times to $\vx_1^m, \vx_2^m, \ldots, \vx_T^m$ following the Markov process below:
\begin{equation}
\vspace{-2pt}
p(\vx_t^m | \vx_{t-1}^m)= \mathcal{N}(\vx^m_t;\sqrt{1-\beta_t}\vx^m_{t-1}, \beta_t \mathbf{I})\,,
\end{equation}
where $t \in [1, 2, \ldots, T]$ denotes the timestep and $\beta_{1:T}$ is the variance schedule of noise, held as hyper-parameters.

Thanks to the properties of Gaussian distribution, we can directly sample $\vx_t^m$ without the recursive formulation by:
\begin{equation}
\vspace{-2pt}
p(\vx_t^m | \vx_0^m)= \mathcal{N}(\vx^m_t;\sqrt{\bar{\alpha}_t}\vx^m_0, (1 - \bar{\alpha}_t) \mathbf{I})\,,
\end{equation}
which can be reparameterized to $\vx_t^m$\,$=$\,$\sqrt{\bar{\alpha}_t} \vx_0^m + \sqrt{1 - \bar{\alpha}_t}\rvepsilon$ for $\rvepsilon$\,$\sim$\,$\mathcal{N}(\mathbf{0}, \mathbf{I})$. $\alpha_t$\,$=$\,$1$\,$-$\,$\beta_t$ and $\bar{\alpha}_t$\,$=$\,$\prod_{i=1}^t \alpha_i$. The variance schedule makes sure that $\bar{\alpha}_T$ of the final timestep $T$ is sufficiently small so that $p(\vx_T^m)$, the end of the forward process, approximates the standard normal distribution~$\mathcal{N}(\mathbf{0}, \mathbf{I})$ well, which is the starting point of the reverse process. 

We want to generate masked regions by sampling from $p(\vx_0^m | \vx_0^v)$, which is approximated by recursively sampling from $p(\vx_{t-1}^m | \vx_t^m, \vx_0^v)$ starting from $\vx_T^m$\,$\sim$\,$\mathcal{N}(\mathbf{0}, \mathbf{I})$. When the variance of noise $\beta_t$ in each step $t$ is small enough, $p(\vx_{t-1}^m | \vx_t^m, \vx_0^v)$ can also be also considered Gaussian distributed~\cite{sohl2015deep}, to be approximated by a deep network. We optimize the \textit{simple} objective proposed by DDPM~\cite{ddpm}:
\begin{equation}
\vspace{-2pt}
\mathcal{L}_{\mathrm{simple}}=\mathbb{E}_{t, \vx_0, \mathbf{\rvepsilon}}\parallel \vx_0^m - D_{\theta}(\vx_t^m, t, E_{\phi}(\vx_0^v))\parallel^2\,.
\end{equation}
This mean squared error objective is simplified from the variational bound. Unlike DDPM using $\mathbf{\rvepsilon}$-prediction, which predicts the noise, our model predicts the denoised masked region $\vx_0^m$. These two formulations are interchangeable, and both are commonly adopted in diffusion models~\cite{dalle-2}. $E_{\phi}$ is the encoder, which projects the visible image content $\vx_0^v$ into the latent space. $D_{\theta}$ is the decoder predicting denoised input, from the noisy $\vx_t^m$, the timestep $t$, and the visible latent $E_{\phi}(\vx_0^v)$.

\subsection{Architecture Design}
We instantiate the above conditional diffusion model following the asymmetric design of MAE, introducing no extra training cost. The overall architecture is built solely on Vision Transformers (ViT)~\cite{vit}, which is in contrast to the typical choice of U-Net~\cite{unet} backbones in diffusion models~\cite{ddpm, adm, dalle-2, imagen}. In this way, we enable a straightforward evaluation of \ourmodel's ability on downstream recognition tasks and an apple-to-apple comparison to other self-supervised pre-training methods.

\paragraph{Encoder.} The encoder takes a standard ViT. Specifically, the training images are first divided into non-overlapping patches, among which most are selected as the visible patches $\vx_0^v$ and the others are the masked $\vx_0^m$. The ViT encoder $E_\phi(\cdot)$ only operates on the visible patches and encodes each of these patches into the latent space. The encoded $E_\phi(\vx_0^v)$ then serves as the condition of the generation task performed by the decoder, providing hints of the masked object. After the pre-training stage, only the encoder is fine-tuned to downstream tasks.

\paragraph{Decoder.}
\label{sec:decoder} The decoder takes in noisy masked patches $\vx_t^m$ as the input. The noise levels of these patches, denoted by the timestep $t$, are integers uniformly sampled in $[1,T]$ during training. As in ViT, we first project these noisy patches to noise tokens using a linear layer. The timestep $t$ can be specified by adding sinusoidal embeddings to the noise tokens together with the positional embeddings. However, our experiments show that the addition of the $t$ embedding or the lack thereof does not make a big difference to both downstream recognition and inpainting generation, which suggests that the decoder can automatically determine the level $t$ of noisy patches when conditioned on clean patches.

We explore three different decoder configurations, which differ in how the attention modules are applied to the visible latents and the noise tokens:

\begin{figure}[!t]
\centering
\includegraphics[width=1.0\linewidth]{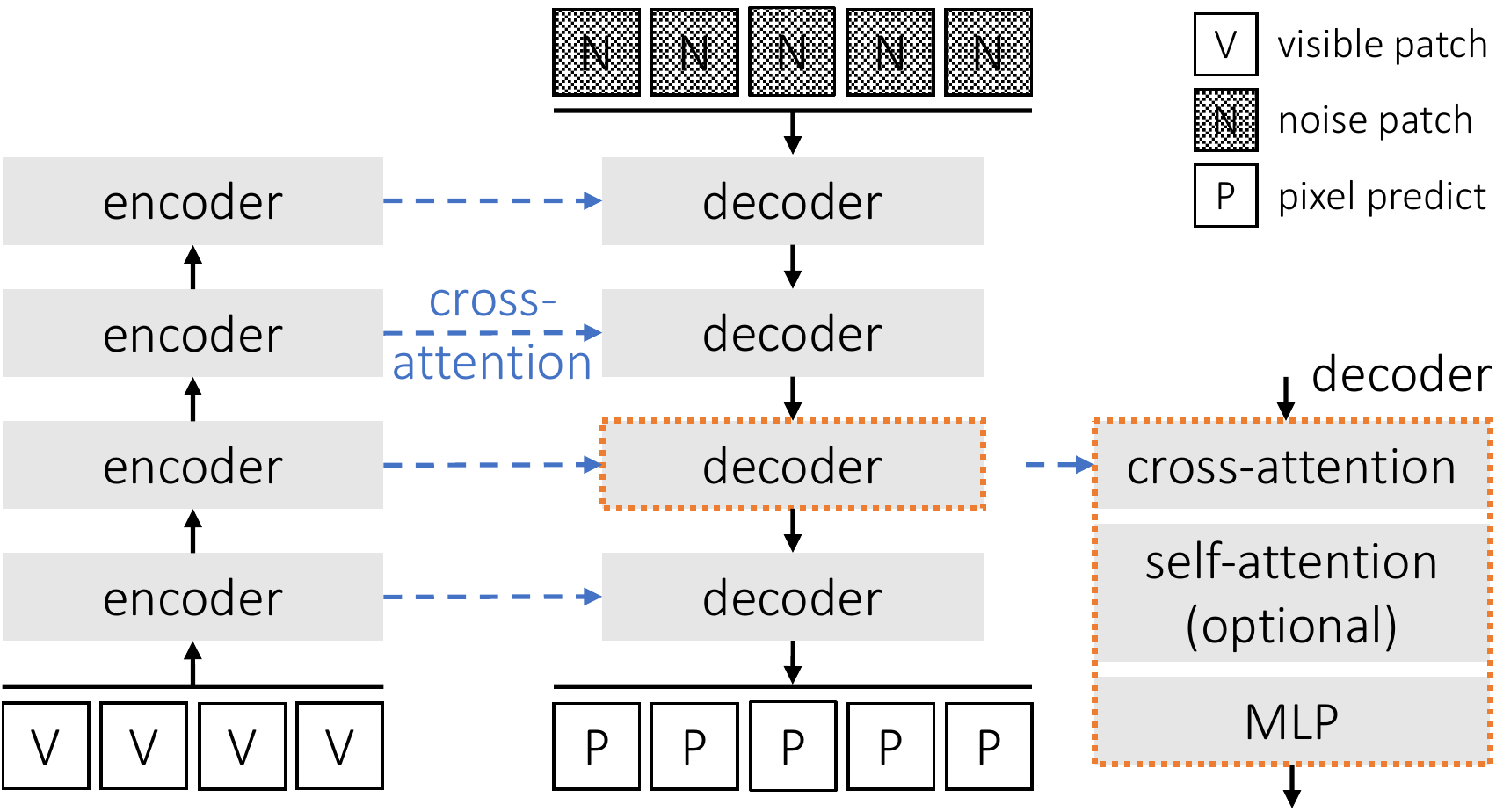}
\caption{\textbf{\ourmodel with cross-self and cross decoder.} The decoder inputs are noisy patches at randomly sampled noise levels. The decoder outputs are the predicted pixels. The decoder blocks are top-down connected to the encoder blocks. Cross-self decoder has self-attention in each block, while cross decoder does not.}
\label{fig:decoder}
\vspace{-10pt}
\end{figure}

\begin{itemize}[leftmargin=*]
\setlength\itemsep{0.0em}
\item \emph{Joint decoder} applies Transformer blocks on the concatenated sequence of the visible latents from the last encoder block and the noise tokens. Each Transformer block operates equally on the visible latents and the noise tokens, with a self-attention layer followed by an MLP.
\item \emph{Cross-self decoder} resembles the encoder-decoder design in the original Transformer~\cite{transformer}. In each decoder block, the noise tokens first attend to the visible latents with a \textit{cross}-attention layer and then attend to each other noise tokens with a \textit{self}-attention layer. The cross-attention enables the decoder to attend to the visible latents of different encoder blocks, rather than solely to the final block as in joint decoder. In this way, the encoder and the decoder can be connected in a U-shaped manner illustrated in \cref{fig:decoder}, which is typically advantageous for image generation and dense predictions such as U-Net~\cite{unet}.
\item \emph{Cross decoder} is similar to cross-self decoder, but excludes the self-attentions between the noise tokens. In other words, each noise token independently attends to the visible latents from the encoder in a top-down manner as in \cref{fig:decoder}, \textit{not} being aware of the existence of other noise tokens. With a much smaller attention map, cross decoder is the most efficient among the three. Specifically, training with cross decoder is overall {\raisebox{0.5ex}{\texttildelow}}15\% more efficient than joint decoder, using a ViT-L encoder and an eight-block decoder of width 512.
\end{itemize}
\noindent We ablate the three decoder architectures in \cref{sec:decoder-representation,sec:dec-generation}.

\paragraph{Sampling.} For inference, the encoder $E_{\phi}(\vx_0^v)$ forwards the visible patches \textit{only once} while the decoder unfolds from Gaussian noise to the sampled image for $T$ times iteratively following DDPM~\cite{ddpm}. As illustrated in \cref{fig:main-viz}, the predictions gradually become less blurry as the iterative unfolding progresses, and both small-scale structures of the objects (\eg, the mouth of the monkey) and high-frequency details (\eg, the fur of the fox) emerge in the generated images. After $T$ iterations, we obtain high-quality samples at $t$\,$=$\,$0$. 

\paragraph{CLIP target.} To compare to recent prior works that use CLIP~\cite{clip}, we also explore a version for predicting CLIP features together with the original pixel prediction task by simply using a second decoder. The prediction is optimized by minimizing the cosine distance to the CLIP features of the masked patches, similar to MaskFeat~\cite{maskfeat} and MILAN~\cite{milan}. As we show in \cref{tab:clip}, predicting CLIP features not only enhances recognition ability but also brings improved inpainting generation quality. 

\paragraph{Video.} The above model are described in the context of \textit{image} operations. However, it can be straightforwardly extended to the spatiotemporal domain of \textit{videos}. Most operations are shared, except that there is an additional temporal dimension. Each token now represents a space-time cube corrupted by noise. The masking is performed by randomly sampling space-time cubes. The prediction target are pixels of a single time slice of each masked space-time cube.

\subsection{Connection between Diffusion and MAE} 
We discuss the connection between diffusion models and MAE. The two are different at first glance: They have different purpose (image generation \vs self-supervised pre-training), inputs (noisy images \vs masked images), outputs (noise \vs pixels) and architectures (U-Net \vs ViT). Consider \cref{fig:main-viz}, at the first timestep $t$\,$=$\,$1000$ where the inputs of the decoder approximate Gaussian noise, the pixel predictions of our model are visually similar to the predictions of MAE, both capturing only the blurry and rough structures of the objects. Moreover, \ourmodel trained with only the $t$\,$=$\,$1000$ noise level obtains a similar fine-tuning accuracy of around 85.0\% as MAE, shown in \cref{fig:decoder-noise}. This consistently observed similarity makes sense because both the Gaussian noise in \ourmodel at $t$\,$=$\,$1000$ and the learnable mask token in MAE contain no image signal.

This observation suggests a close relationship between MAE and diffusion models, that MAE effectively performs the first inference step ($t$\,$=$\,$1000$) of diffusion models. On the other hand, the other steps of diffusion models generate more and more intricate high-frequency details, which is lacking in the predictions of MAE. Overall, the connection suggests that MAE can be viewed as a single-step patch-conditioned diffusion model, and that the success of MAE in downstream recognition tasks is \textit{in line with} the philosophy of generative pre-training. Correspondingly, diffusion models, the first step of whom effectively performs MAE, are potentially good recognition models.

\section{Empirical Study}
\label{sec:empirical}

\paragraph{Settings.} We pre-train our \ourmodel on the IN-1K~\cite{imagenet} training set. If not specified, the encoder is a vanilla ViT-L~\cite{vit} without any modification, and the decoder is of depth 8 and width 512. The eight cross-(self) decoder blocks spread top-down uniformly attending to the outputs of the 24 encoder blocks of ViT-L. The data augmentation is randomly resized cropping, and the masking strategy is to mask out 75\% patches randomly. All models are pre-trained for 400 epochs. For the diffusion setup, the variance follows a linear schedule~\cite{ddpm}, and the number of timesteps $T$ is set to 1000. Details 
 are in the Appendix.

From the perspective of pre-training for \textit{recognition}, we report end-to-end fine-tuning top-1 accuracy on the IN-1K validation set. The model is both pre-trained and fine-tuned at 224$^2$ resolution. By default, we use our cross decoder design, and the prediction targets are per-patch normalized.

For the task \textit{generative} inpainting, we report the common LPIPS$\downarrow$~\cite{lpips}, which is a learned distance metric based on the deep feature space of AlexNet~\cite{alexnet}. Following the setting of RePaint~\cite{repaint}, we compute LPIPS$\downarrow$ over 100 images from the IN-1K validation set, and train and evaluate the model with image size 256$^2$. We use cross-self decoder, and the prediction targets are unnormalized pixels.

\begin{figure}[!t]
\centering
\includegraphics[width=\linewidth]{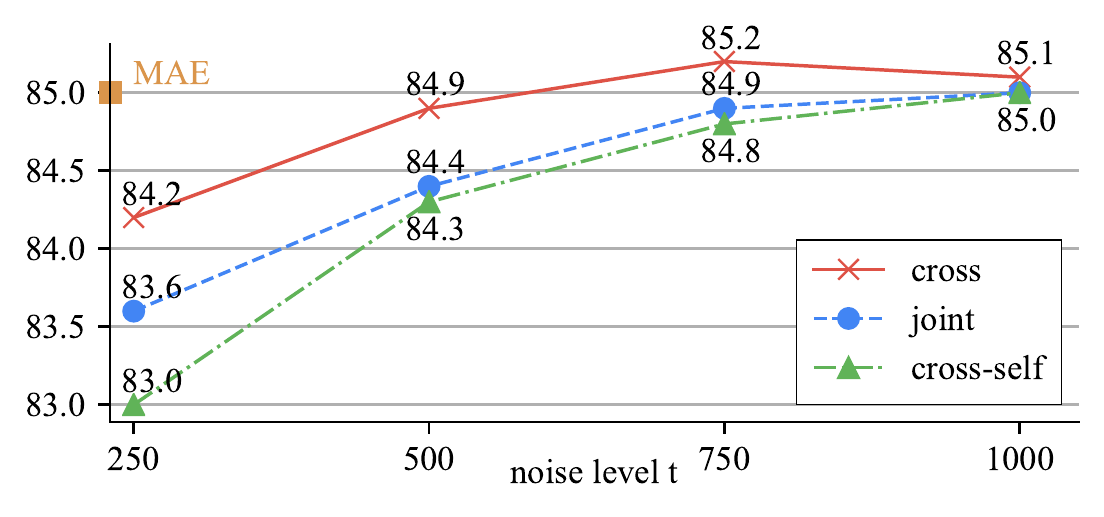}
\caption{\textbf{Comparison of different decoder architectures} across noise levels (\textit{x-axis}) by fine-tuning top-1 (\%) on IN-1K val set (\textit{y-axis}). Each \ourmodel model is pre-trained with a \textit{single} noise level. All models are ViT-L pre-trained for 400 epochs.}
\label{fig:decoder-noise}
\vspace{-10pt}
\end{figure}

\paragraph{Decoder architecture -- classification.}
\label{sec:decoder-representation}
We find that the three decoder architectures impact fine-tuning for downstream classification differently across different noise levels. To demonstrate this, we train the models at various \textit{single} noise levels, analyzed in \cref{fig:decoder-noise}. A smaller $t$ corresponds to less noise while a larger $t$ denotes more noise. At $t$\,$=$\,$1000$, the inputs have almost no image signals but noise. At $t=0$, the inputs would be the clean image patches.

First, the fine-tuning accuracy drops for all three decoders when $t$ is lower. At a small noise level $t$, the disruption to image signals by noise is less severe. Consequently, the decoder needs less conditioning on the visible latents to predict image signals, which weakens the pre-training of the encoder. On the other hand, the model is now tasked to generate more low-level, high-frequency details, which are considered to be less useful for downstream recognition.

Second, the drop rates of joint and cross-self decoder are faster than that of cross decoder. This is because both joint decoder and cross-self decoder have a shortcut to exploit the image signals of the neighboring noisy patches and bypass the visible latents of the encoder. This shortcut is amplified especially when $t$ is small. In contrast, cross decoder, whose inputs do not see other noisy patches, avoids this shortcut.

In summary, our experiments show that the pre-training performance is closely related to the difficulty of the denoising task. Nonetheless, the pre-training helps recognition in all cases compared to its from-scratch counterpart of 82.6\% top-1~\cite{mae}. We make cross decoder the default option for pre-training for downstream recognition tasks.

\begin{figure}[!t]
\centering
\includegraphics[width=1.0\linewidth]{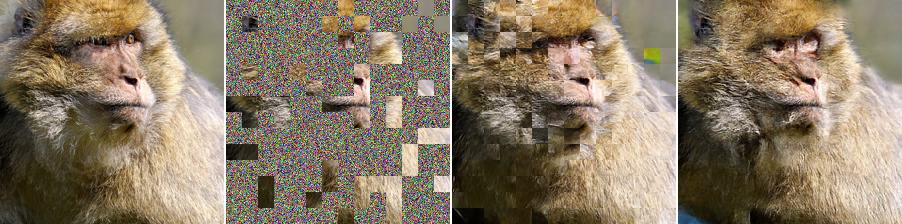}

\tablestyle{0pt}{0.0}
\begin{tabular}{x{59}x{59}x{59}x{59}}
ground-truth & input & cross & cross-self
\end{tabular}
\caption{\textbf{Generations of cross and cross-self decoders.} The generation of cross decoder is patch-wise discontinuous, while cross-self decoder generates visually continuous samples.}
\label{fig:decoder-synthesis}
\vspace{-10pt}
\end{figure}

\paragraph{Decoder architecture -- inpainting.}
\label{sec:dec-generation}
Next, we discuss the effect of different decoders on inpainting. For the cross decoder, the noise tokens are independent of each other and do not see other noise tokens. Therefore, the generated image is patch-wise discontinuous. Even so, these patches still make sense semantically at their positions, show in \cref{fig:decoder-synthesis}.

The generations of both joint and cross-self decoders are continuous and visually pleasant. Comparing these two, cross-self decoder achieves a better LPIPS$\downarrow$ score of 0.216 on the evaluation set while joint decoder obtains 0.224. We credit the better generation of cross-self decoder to its U-shape design following U-Net, which is commonly suitable for dense predictions. We make cross-self decoder the default option for inpainting.

\begin{table}[!h]
\centering
\tablestyle{8pt}{1.02}
\begin{tabular}{z{50}x{30}|x{50}}
 range of $t$ & $\rho$ & top-1 \\
 \shline
 1000 & 1.0 & 85.1 \\
 0{\raisebox{0.5ex}{\texttildelow}}1000 & 1.0 & 84.7 \\
 \hline
 250{\raisebox{0.5ex}{\texttildelow}}1000  & 1.0 & 84.7 \\
 500{\raisebox{0.5ex}{\texttildelow}}1000  & 1.0 & 85.1 \\
 \hline
 0{\raisebox{0.5ex}{\texttildelow}}1000  & 0.8 & 85.1 \\
 0{\raisebox{0.5ex}{\texttildelow}}1000  & 0.6 & 85.0 \\
\end{tabular}
\caption{\textbf{Noise schedule.} We first vary the range of $t$ with different starting points. We then modify the default linear schedule with a hyper-parameter $\rho$ as the exponent to each variance $\beta_t$.}
\label{tab:noise-schedule} 
\vspace{-10pt}
\end{table}

\paragraph{Noise variance schedule.}
\label{sec:noise-schedule}
Next, we train our model with mixed noise levels to complete the diffusion process in \cref{tab:noise-schedule}. We use the linear schedule following DDPM~\cite{ddpm} by default, where the variances $\beta_{[1:T]}$ increase linearly.

We first vary the range of $t$, starting from $t$\,$=$\,$250$ and $t$\,$=$\,$500$. Similar to our observations on training with single noise levels, the less the $t$ range includes lightly noisy samples, the better the fine-tuning accuracy is. When $t$ is mixed from $500$ to $1000$, the pre-trained model is just as good as the case of training with only $t$\,$=$\,$1000$ samples.

We then directly modify the linear schedule with a hyper-parameter $\rho$, which exponentiates the variances $\beta_t$ to $\beta_t^{\rho}$. When $\rho$ is smaller than one, it enlarges each variance $\beta_t$, therefore amplifying the amount of noise at each timestep. Consequently, the fine-tuning top-1 improves with a smaller $\rho$ from 84.7\% to around 85.0\%. However, the gain on downstream classification is at the cost of a drop in the inpainting generation. Specifically, LPIPS$\downarrow$ adversely increases from 0.216 with the default linear schedule ($\rho$\,$=$\,$1.0$) to 0.228 with $\rho$\,$=$\,$0.6$. This suggests another trade-off between pre-training for recognition and inpainting generation. We detail $\rho$ formulation in the Appendix.

\paragraph{Prediction target.} We next study the influence of prediction target in \cref{tab:target}. Both noise and pixel prediction are commonly adopted in generative diffusion models~\cite{ddpm, dalle-2}. However, when evaluated for downstream classification, only predicting the pixels can serve as the target, obtaining 84.3\% top-1. The noise prediction, however, can not be stably fine-tuned for the downstream classification task. Regarding the pixel prediction, the per-patch normalization~\cite{mae} consistently helps the fine-tuning accuracy.

\begin{table}[!t]
\centering
\tablestyle{8pt}{1.02}
\begin{tabular}{x{35}|x{35}x{35}x{50}}
target & noise & pixel & pixel w/ norm \\
\shline
top-1  & unstable & 84.3 & 85.1   \\
\end{tabular}
\vspace{4pt}
\caption{\textbf{Prediction targets} of noise, pixels, and per-patch normalized pixels. The noise entry is unstable in fine-tuning.}
\label{tab:target}
\vspace{-10pt}
\end{table}

\begin{figure}[!h]
\centering
\includegraphics[width=\linewidth]{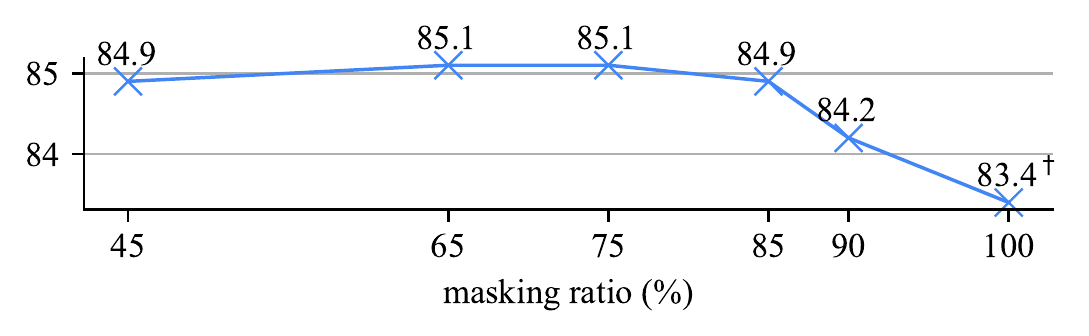}
\vspace{-17pt}
\caption{\textbf{Masking ratio.} Varying the masking ratio of random patch masking. $^\dagger$For 100\% masking, we use the encoder-only architecture instead of the asymmetric encoder-decoder architecture.}
\label{fig:masking-ratio}
\vspace{-5pt}
\end{figure}

\paragraph{Masking ratio.} We consider different masking ratios for the random masking strategy. We report the fine-tuning accuracy in \cref{fig:masking-ratio}. We observe that accuracy is stable with masking ratios from 45\% to 85\% and has a drop when the masking ratio increases to 90\%. When the masking ratio is increased to 100\%, \ie, no masking at all, the fine-tuning top-1 drops to 83.4\%. This suggests the importance of visible patch conditioning for \ourmodel pre-training. On the other hand, the recognition performance is still enhanced compared to its from-scratch baseline of 82.6\% top-1~\cite{mae}.

\begin{table}[!h]
\centering
\tablestyle{3pt}{0.0}
\begin{tabular}{cc}
\tablestyle{3pt}{1.02}
\begin{tabular}{x{19}|x{28}x{20}}
 mask       & random & center\\
\shline
top-1      & 85.1 & 84.3 \\
\end{tabular} & 
\tablestyle{3pt}{1.02}
\begin{tabular}{x{67}|x{29}x{20}}
training masking & random & center \\
\shline
center LPIPS$\downarrow$   & 0.142 & 0.125 \\
\end{tabular}
\end{tabular}
\vspace{-2pt}
\caption{\textbf{Center masking.} The left reports the fine-tuning top-1. The right reports LPIPS$\downarrow$ of the two models trained with random and center mask, respectively, but evaluated \textit{both with center mask}.}
\label{tab:masking-strategy}
\vspace{-10pt}
\end{table}

\begin{figure}[!t]
\tablestyle{0pt}{0.0}
\begin{tabular}{cc}
\includegraphics[width=0.5\linewidth]{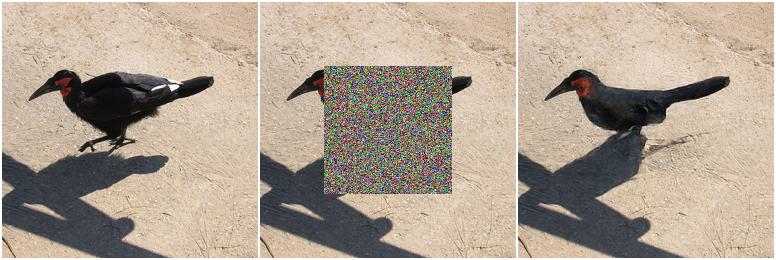} & \includegraphics[width=0.5\linewidth]{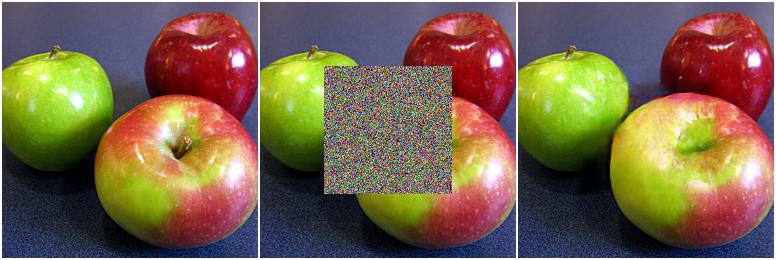} \\
\includegraphics[width=0.5\linewidth]{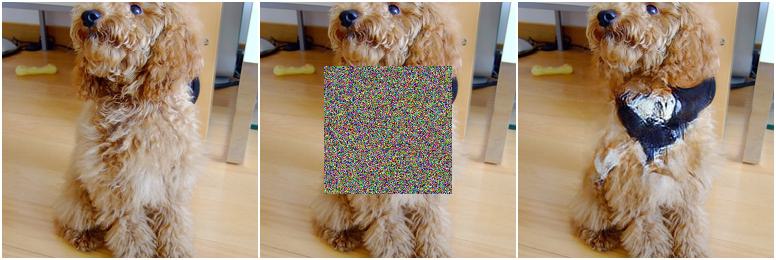} & \includegraphics[width=0.5\linewidth]{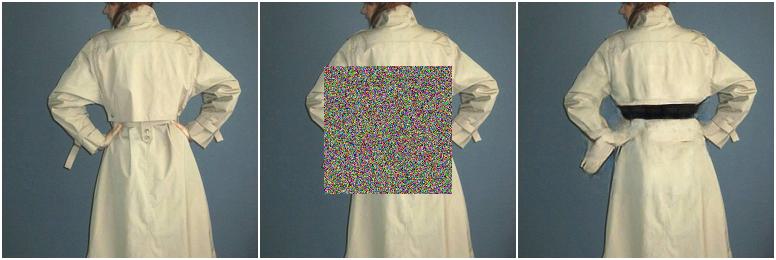} \\
\end{tabular}

\tablestyle{0pt}{0.0}
\begin{tabular}{x{39}x{39}x{40}x{39}x{39}x{40}}
original & input & sample & original & input & sample
\end{tabular}

\caption{\textbf{Samples of center masking inpainting} of \ourmodel. For a 256$^2$ image, we mask out the center 128$^2$ block.}
\label{fig:center}
\vspace{-5pt}
\end{figure}

\paragraph{Center masking.} We now compare random masking with  a more common evaluation protocol for inpainting tasks, center masking, which block-wise masks out the center 25\% region of images as in \cref{fig:center}. We first train with two different maskings and then evaluate both models by fine-tuning top-1. As shown in \cref{tab:masking-strategy} (\textit{left}), random 75\% mask works better for fine-tuning.

We then evaluate \textit{both random and center mask} trained models by inpainting LPIPS$\downarrow$ with \textit{center mask} in \cref{tab:masking-strategy} (\textit{right}), the common inpainting protocol. The random mask trained model provides reasonable center block inpainting, reporting 0.142 LPIPS$\downarrow$ on the evaluation set. This shows that the model still obtains block-wise inpainting ability although trained with random masks. When trained with center mask, the inpainting quality is further improved, and LPIPS$\downarrow$ favorably decreases to 0.125.

\begin{table}[!h]
\centering
\tablestyle{5.5pt}{1.00}
\begin{tabular}{cc}
\tablestyle{4pt}{1.02}
\begin{tabular}{c|x{30}|x{30}}
depth & top-1 & LPIPS$\downarrow$ \\
\shline
4  & 85.0 & 0.220 \\
8  & 85.1 & 0.216 \\
12 & 85.1 & 0.215 \\
\end{tabular} & 
\tablestyle{4pt}{1.02}
\begin{tabular}{c|x{30}|x{30}}
width & top-1 & LPIPS$\downarrow$ \\
\shline
256  & 85.0 & 0.239 \\
512  & 85.1 & 0.216 \\
1024 & 84.9 & 0.208 \\
\end{tabular}
\end{tabular}
\caption{\textbf{Decoder size.} The left table varies the depth of the decoder with the default width of 512. The right table varies the width of the decoder with the default depth of eight blocks.}
\label{tab:decoder-size}
\vspace{-10pt}
\end{table}

\paragraph{Decoder size.} We discuss the effect of the decoder size in \cref{tab:decoder-size}. By default, we use eight decoder blocks, each of which has 512 channels. We do not observe significant differences in the fine-tuning performance when varying the decoder depth or the width, all of which have around 85.0\% top-1. For inpainting, we observe a notable drop if the decoder size is reduced and a gain when the decoder size is enlarged. With only four decoder blocks, the LPIPS$\downarrow$ adversely increases from 0.216 to 0.220. A narrower decoder of width 256 leads to an even worse LPIPS$\downarrow$ of 0.239. Using an extra wide decoder with 1024 channels, the inpainting is largely improved to 0.208 LPIPS$\downarrow$.

\begin{table}[!h]
\vspace{-5pt}
\tablestyle{1pt}{1.00}
\begin{tabular}{cc}
\tablestyle{3.5pt}{1.00}
\begin{tabular}{c|ccc}
target & pixel & CLIP & \footnotesize pixel+CLIP \\
\shline
top-1      & 85.1   & 86.5 & 86.7   \\
\end{tabular} & 
\tablestyle{3.5pt}{1.00}
\begin{tabular}{c|cc}
target & pixel & \footnotesize pixel+CLIP \\
\shline
\footnotesize LPIPS$\downarrow$ & 0.226 & 0.216   \\
\end{tabular}
\end{tabular}
\caption{\textbf{CLIP target.} Varying the prediction target between pixel, CLIP feature, and the combination of the two.}
\vspace{-10pt}
\label{tab:clip}
\end{table}

\paragraph{CLIP target.}
\label{sec:clip}
Boosting pre-training with CLIP~\cite{clip}, a vision-language model trained on a 400M in-house dataset, has been popular driven by its strong performance in downstream tasks. MaskFeat~\cite{maskfeat} first showed improved performance for masked feature prediction. Then algorithms utilizing CLIP either regress CLIP features~\cite{mvp,milan}, or predict CLIP-aided tokens~\cite{beitv2}. In this work, we use a second cross decoder to regress CLIP features alongside the original pixel decoder, multitasking on both masked pixel and CLIP feature prediction. In \cref{tab:clip} (\textit{left}), we first compare the options by fine-tuning. Solely regressing CLIP features achieves a high accuracy 86.5\%, compared to the default pixel predicting baseline 85.1\%. Multitasking on both targets further improves the accuracy to 86.7\%.

Beyond the improvement on fine-tuning, we observe in \cref{tab:clip} (\textit{right}) that combining CLIP targets can also improve inpainting. The LPIPS$\downarrow$ decreases from 0.226 to 0.216 when multitasking with CLIP regression. This suggests that a better semantic understanding, in this case by learning from CLIP, can notably help image inpainting quality.

\paragraph{Discussion.} We have thoroughly studied the design aspects of \ourmodel in terms of downstream classification and generative inpainting performance. Our experiments show that, for many aspects, pre-training for downstream classification and generative inpainting do not share the optimal setting, \eg, the decoder architecture and the noise variance schedule. For other aspects, settings favoring inpainting quality do not impact pre-training ability, \eg, decoder size. Interestingly, using CLIP benefits both downstream classification and generative inpainting. We use cross decoder, per-patch normalization, and noisy variance schedule for fine-tuning evaluation, and we use cross-self decoder with 1024 width, the default linear schedule, but not per-patch normalization for inpainting evaluation. While the inpainting-oriented model can still improve from-scratch recognition accuracy, we still would find it desirable to have an identical setting that is optimal for both tasks.

\section{Comparison to Prior Work}
\label{sec:experiments}

\begin{table}[t!]
\centering
\tablestyle{2pt}{1.02}
\begin{tabular}{y{70}x{36}x{36}x{36}x{36}}
pre-train                                & w/ CLIP           & ViT-B          & ViT-L          & ViT-H \\
\shline
from-scratch~\cite{mae}                  & \x                & 82.3           & 82.6           & 83.1  \\
\hline
MoCo v3~\cite{mocov3}                    & \x                & 83.2           & 84.1           & -     \\
DINO~\cite{dino}                         & \x                & 82.8           & -              & -     \\
iBOT~\cite{ibot}                         & \x                & 84.0           & 84.8           & -     \\
BEiT~\cite{beit}                         & \x                & 83.2           & 85.2           & -     \\
MaskFeat~\cite{maskfeat}                 & \x                & 84.0           & 85.7           & -     \\
MAE~\cite{mae}                           & \x                & 83.6           & 85.9           & \textbf{86.9}  \\ 
\rowcolor{defaultcolor}
\ourmodel                                & \x                & 83.9           & 85.8           & \textbf{86.9}  \\ 
\shline
MVP~\cite{mvp}                           & \checkmark        & 84.4           & 86.3           & -     \\
MILAN~\cite{milan}                       & \checkmark        & 85.4           & 86.7           & -     \\
BEiTv2$^\dagger$~\cite{beitv2}           & \checkmark        & 85.5           & 87.3           & -     \\
\rowcolor{defaultcolor}
\ourmodel                                & \checkmark        & 84.9           & 86.9           & \textbf{88.0}  \\
\end{tabular}
\vspace{5pt}
\caption{\textbf{Comparison to prior works on IN-1K fine-tuning.} All entries are pre-trained on IN-1K train split at image size 224$^2$. When using CLIP, all ViT-B and ViT-L utilize CLIP ViT-B/16 and ViT-H  uses CLIP ViT-L/14. $^\dagger$BEiTv2's ViT use layer scale~\cite{layerscale} and relative positional embedding~\cite{relpos}.
}
\label{tab:finetuning}
\vspace{-5pt}
\end{table}

\paragraph{ImageNet fine-tuning.} We compare \ourmodel to prior works in fine-tuning for IN-1K classification in \cref{tab:finetuning}.

The first section presents methods that are solely trained on IN-1K training set without access to CLIP models. We pre-train \ourmodel for 1600 epochs here while 400 were used for ablations in \cref{sec:empirical}.
We observe that fine-tuning \ourmodel significantly boosts the scratch accuracy for all ViT-B (+1.6\%), ViT-L (+3.2\%) and ViT-H (+3.8\%) models. This suggests that \ourmodel can provide strong representations for recognition compared to prior generative models (\textit{cf.}~\cref{tab:gen-models}), and is scalable to large models. Compared to prior self-supervised learning works, it outperforms contrastive methods like MoCo v3 and DINO and is comparable masked autoencoders, while being able to \textit{generatively inpaint images with high quality} at the same time.

The second section presents representative works that use CLIP to enhance representation learning. We pre-train our \ourmodel for 800 epochs multitasking in masked pixel and CLIP feature prediction. For ViT-B and ViT-L models, all entries utilize CLIP ViT-B/16. For ViT-H, we use CLIP ViT-L/14. The CLIP models are pre-trained on in-house WIT400M~\cite{clip} and lift accuracy for both \ourmodel and other works. Combined with CLIP, \ourmodel obtains a strong IN-1K top-1 of 88\% with ViT-H, while ViT-B lags behind because of its insufficient capacity for multitasking.

\begin{figure}[!t]
\vspace{-7pt}
\centering
\tablestyle{0.3pt}{0.2}
\begin{tabular}{ccccc}
\includegraphics[width=0.2\linewidth]{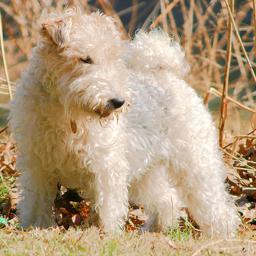} & 
\includegraphics[width=0.2\linewidth]{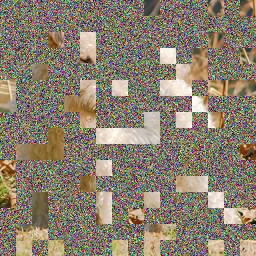} & 
\includegraphics[width=0.2\linewidth]{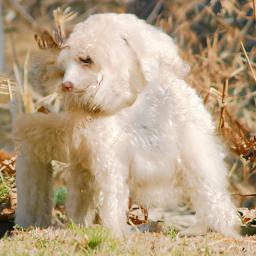} & 
\includegraphics[width=0.2\linewidth]{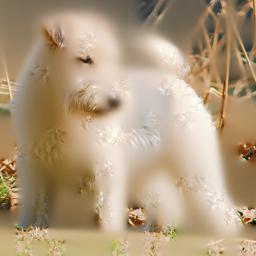} &
\includegraphics[width=0.2\linewidth]{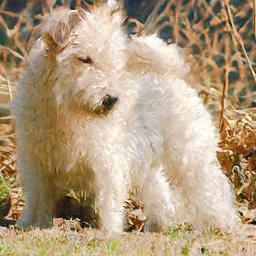} \\

\includegraphics[width=0.2\linewidth]{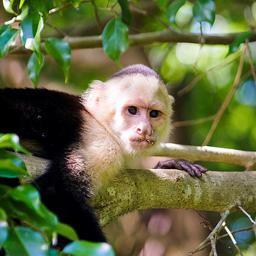} & 
\includegraphics[width=0.2\linewidth]{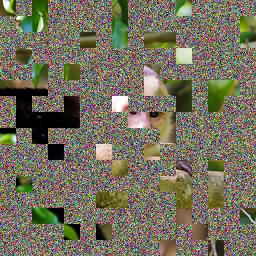} & 
\includegraphics[width=0.2\linewidth]{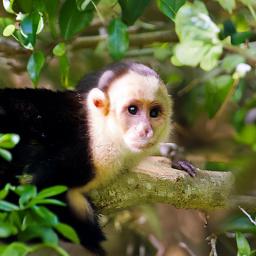} & 
\includegraphics[width=0.2\linewidth]{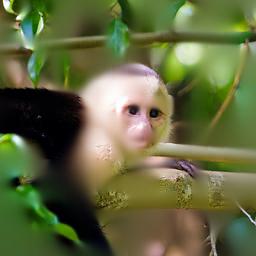} &
\includegraphics[width=0.2\linewidth]{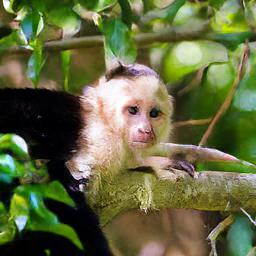} \\

\includegraphics[width=0.2\linewidth]{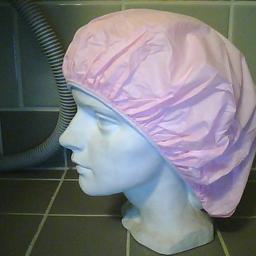} & 
\includegraphics[width=0.2\linewidth]{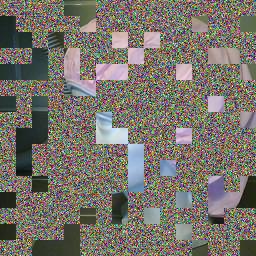} & 
\includegraphics[width=0.2\linewidth]{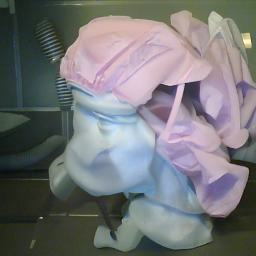} & 
\includegraphics[width=0.2\linewidth]{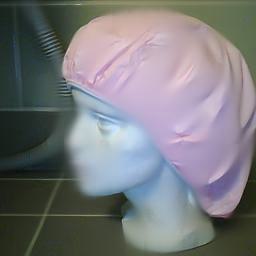} &
\includegraphics[width=0.2\linewidth]{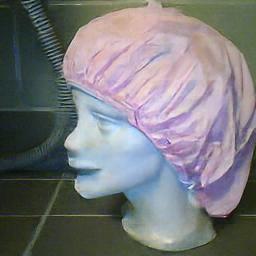} \\
\end{tabular}

\tablestyle{0pt}{1.2}
\begin{tabular}{x{48}y{48}x{48}x{48}x{47}}
original & 75\% random &  RePaint & MAE & \ourmodel \\
\end{tabular}
\caption{\textbf{Visualizations on inpainting} of three different algorithms. Generations of \ourmodel have finer details than MAE \textit{and} are more semantically meaningful than RePaint.}
\label{fig:repaint}
\vspace{-15pt}
\end{figure}

\begin{table}[!h]
\vspace{-10pt}
\centering
\tablestyle{9pt}{1.0}
\begin{tabular}{llll}
arch. & pre-train  & AP$^\textrm{box}$ & AP$^\textrm{mask}$ \\
\shline
Swin-L & Sup., IN-21K   & 52.4              & 46.2              \\
ViT-L  & MAE, IN-1K    & 55.6              & 49.2              \\
ViT-L  & DiffMAE, IN-1K & 55.3              & 49.0              \\
\end{tabular}
\caption{\textbf{Comparisons on COCO instance segmentation.}}
\label{tab:coco}
\vspace{-15pt}
\end{table}

\paragraph{COCO instance segmentation.} We conduct COCO~\cite{coco} instance segmentation with ViTDet~\cite{vitdet} using Mask R-CNN~\cite{maskrcnn} in \cref{tab:coco}. Our results are pre-trained on IN-1K and slightly lower than those of MAE, while being much stronger than \eg supervised Swin-L trained on IN-21K.

\begin{table}[t!]
\vspace{-8pt}
\hspace*{-7pt}
\centering
\tablestyle{1.0pt}{1.04}
\resizebox{1.0\linewidth}{!}{
\begin{tabular}{l|l|c|c|c|r|l|r}
model  & pre-train  & \footnotesize extra data & top-1 & top-5 & input size & FLOPs &  \scriptsize{param}\\
\shline \hline
SlowFast~\cite{slowfast} & scratch & - & 79.8 & 93.9 & 64\x224$^2$ & 234\x3\x10   & 60  \\
X3D-XL~\cite{x3d}  & scratch & - & 79.1 & 93.9 & 16\x 312$^2$ & 48\x3\x10  & 11  \\ 
MViT-B~\cite{mvit} & scratch & - & 81.2 & 95.1 & 64\x224$^2$ & 455\x3\x3 & 37  \\
MViTv2-B~\cite{mvitv2} & scratch & - & 82.9 & 95.7 & 32\x224$^2$ & 255\x1\x5 & 51  \\
\hline
MViTv2-L~\cite{mvitv2} & MaskFeat~\cite{maskfeat} & - & 84.3 & 96.3 & 16\x224$^2$ & 377\x1\x10 & 218 \\
ViT-L~\cite{vit} & MAE~\cite{videomae} & - & 84.8 & 96.2 & 16\x224$^2$ & 598\x3\x7 & 304 \\ 
\rowcolor{defaultcolor}
ViT-L\cite{vit} & \ourmodel & - & 84.5 & 96.3 & 16\x224$^2$ & 598\x3\x7 & 304 \\
\shline
\hline
ViT-L/14~\cite{vit} & EVL~\cite{evl} & \footnotesize WIT400M & 86.9 & 97.4 & 16\x224$^2$ & 1348\x1\x3 & n/a \\
ViT-L/14~\cite{vit} & X-CLIP~\cite{xclip} & \footnotesize WIT400M & 87.1 & 97.6 & 8\x224$^2$ & 658\x4\x3 & n/a \\
\rowcolor{defaultcolor}
ViT-L/14~\cite{vit} & \ourmodel & \footnotesize WIT400M & 87.4 & 97.5 & 16\x224$^2$ & 828\x3\x10 & 304 \\
\hline
ViT-L/14~\cite{vit} & EVL~\cite{evl} & \footnotesize WIT400M & 87.7 & 97.8 & 32\x336$^2$ & 6065\x1\x3 & n/a \\
ViT-L/14~\cite{vit} & X-CLIP~\cite{xclip} & \footnotesize WIT400M & 87.7 & 97.4 & 16\x336$^2$ & 3086\x4\x3 & n/a \\
\rowcolor{defaultcolor}
ViT-L/14~\cite{vit} & \ourmodel & \footnotesize WIT400M & \textbf{88.1} & 97.8  & 32\x$280^2$ & 2588\x4\x3 & 304 \\
\end{tabular}
}
\caption{\textbf{Comparison with previous works on Kinetics-400}. We report the inference cost with a single ``view" (temporal clip with spatial crop) $\times$ the number of views (FLOPs\x view$_\text{space}$\x view$_\text{time}$). Magnitudes are Giga ($10^9$) for FLOPs and Mega ($10^6$) for Param.}
\label{tab:k400-finetune}
\vspace{-7pt}
\end{table}

\paragraph{Video classification.} We compare \ourmodel to prior work on K400 fine-tuning for video classification in \cref{tab:k400-finetune}.

The first section presents results that use no in-house data. Our model largely improves from the from-scratch models and achieves comparable results to other algorithms, which use masked autoencoding methodology for pre-training. In the second section, we present the results that use extra in-house data. Especially, we compare to EVL~\cite{evl} and X-CLIP~\cite{xclip}, both using CLIP ViT-L/14 model, which is trained on the in-house text-image dataset WIT400M~\cite{clip}. \ourmodel obtains 87.4\% top-1 with a regular ViT-L/14 model, outperforming these works. This number is further improved to a high 88.1\% with a larger input size $280^2$ and a longer temporal duration of 32 frames.

\begin{table}[h!]
\centering
\tablestyle{4pt}{1.02}
\begin{tabular}{y{60}y{60}x{40}x{40}}
method                 & arch.                      & random   & center \\
\shline
DSI~\cite{dsi}     & VQ-VAE2~\cite{vqvae2}      & 0.300    & 0.153  \\
RePaint~\cite{repaint} & ADM~\cite{adm}             & 0.303    & 0.160  \\
\ourmodel              & ViT-L                      & 0.208    & 0.125  \\
\ourmodel              & ViT-H                      & 0.205    & 0.121  \\
\end{tabular}
\caption{\textbf{Comparison to prior works on IN-1K \textit{inpainting}} with LPIPS$\downarrow$ for 75\% random mask and 25\% center mask settings. All entries are trained and evaluated at image size 256$^2$.}
\label{tab:inpainting}
\vspace{-17pt}
\end{table}

\paragraph{ImageNet inpainting.} We first qualitatively compare RePaint~\cite{repaint}, MAE, and our \ourmodel on 75\% random mask inpainting in \cref{fig:repaint}. RePaint employs a pre-trained unconditional diffusion model, specifically ADM~\cite{adm}, as the generative prior to perform free-form inpainting. We notice that the generations of RePaint, though with visually pleasant details, often need more precise semantic meaning. In contrast, MAE generates blurry samples. \ourmodel generates visually detailed and semantically meaningful images.

We finally quantitatively compare the inpainting performance in \cref{tab:inpainting}. Both DSI and RePaint are algorithms specially designed for inpainting. We evaluate at image size 256$^2$ following both works. We first compare the algorithms in the 75\% random mask setting where DSI and RePaint's results are obtained by us with their official models. We then compare the three algorithms with the 25\% center mask protocol, which is commonly used by inpainting algorithms. \ourmodel largely improves LPIPS$\downarrow$ on both random 75\% mask and center block mask protocols, setting a new state-of-the-art for ImageNet inpainting. We include more generative inpainting examples in the Appendix.

\section{Conclusion}
\label{sec:conclusion}
We present Diffusion Masked Autoencoders (\ourmodel), a self-supervised framework designed for recognizing and generating images and videos. We accomplish this by integrating masking into diffusion models, thereby transforming conditional diffusion models into masked autoencoders. By re-evaluating the philosophy of generation and viewing it as a tool for genuinely comprehending visual data, we contribute to the current trend in generative models. Ultimately, we aspire for our research to spark further exploration of integrated methods for recognition and generation.

\appendix
\section*{Appendix}
In the Appendix, we first provide implementation details in \cref{sec:impl}, and then provide more qualitative results in \cref{sec:more-viz}.

\section{Implementation Details}
\label{sec:impl}

\subsection{ImageNet Experiments}
\begin{figure}[!h]
\centering
\includegraphics[width=0.75\linewidth]{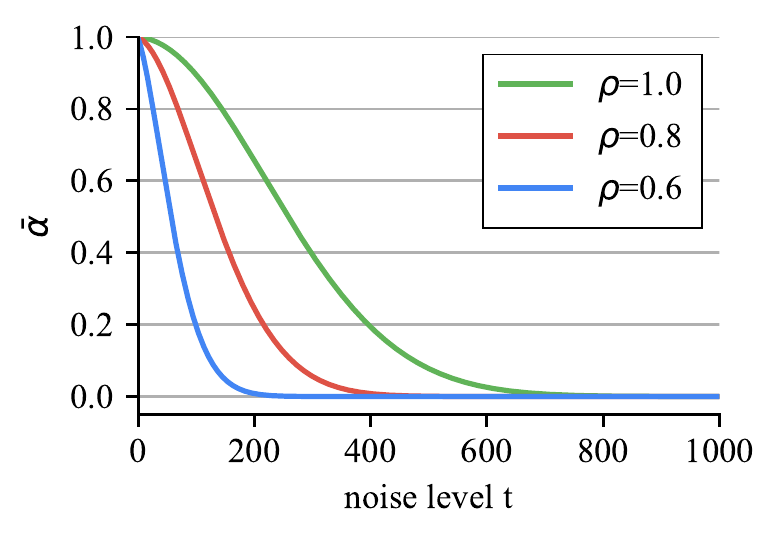}
\vspace{-10pt}
\caption{\small $\bar{\alpha}_t$ throughout the diffusion process with different $\rho$.}
\label{fig:supp_rho}
\vspace{-10pt}
\end{figure}

\paragraph{Noise schedule with $\rho$.} We introduce a hyper-parameter $\rho$ to control the noise level of training inputs. Specifically, we use $\rho$ to exponentiate each variance $\beta_t$ to $\beta_t^\rho$, enlarging these noise variance. Recall that the training samples can be reparameterized to $\vx_t^m$ \,$=$\,$ \sqrt{\bar{\alpha}_t} \vx_0^m + \sqrt{1 - \bar{\alpha}_t}\rvepsilon$, where $\alpha_t$\,$=$\,$1$\,$-$\,$\beta_t$ and $\bar{\alpha}_t$\,$=$\,$\prod_{i=1}^t \alpha_i$. In \cref{fig:supp_rho}, we plot how the values of the data coefficient $\bar{\alpha}_t$ progress with different $\rho$. $\rho$\,$=$\,$1.0$ represents the default linear schedule introduced in DDPM~\cite{ddpm}, where the forward process variances $\beta_t$ increase linearly from $10^{-4}$ to $0.02$. With $\rho$\,$=$\,$0.8$ and $\rho$\,$=$\,$0.6$, the data coefficients $\bar{\alpha}_t$ are lower at each timestep $t$, and the amount of noise is therefore amplified. 

\paragraph{Architecture.}
\label{sec:supp-img-impl}
We use the standard ViT architecture~\cite{vit} in base, large and huge sizes for the encoder. The encoder is followed by Layer Normalization~\cite{ln}. There is a linear projection layer after the layer normalization to match the dimension of the encoder to that of the decoder. We add sinusoidal positional embeddings to both the encoder and decoder inputs for pre-training. We do not use either relative positional embedding~\cite{relpos} or
layer scale~\cite{electra}. The encoder and the decoder use two different linear projections to handle the clean and the noised (masked) inputs, respectively.

\begin{table}[!t]
\tablestyle{3pt}{1.02}
\begin{tabular}{y{80}|x{65}x{65}}
config & ImageNet & Kinetics \\
\shline
optimizer & \multicolumn{2}{c}{AdamW~\cite{adamw}} \\
optimizer momentum & \multicolumn{2}{c}{$\beta_1, \beta_2{=}0.9, 0.95$} \\
weight decay & \multicolumn{2}{c}{0.05} \\
learning rate schedule & \multicolumn{2}{c}{cosine decay~\cite{sgdr}} \\
warmup epochs~\cite{1hour} & \multicolumn{2}{c}{40} \\
augmentation & \multicolumn{2}{c}{hflip, RandomResizedCrop} \\
drop path~\cite{droppath} & \multicolumn{2}{c}{0.0} \\ 
base lr & 1.5e-4 & 8.0e-4 \\
batch size & 4096 & 512 \\
epochs & 1600 & 400 \\
gradient clipping & - & 0.02 \\
repeated aug.~\cite{repeatedaug} & - & 4 \\
\end{tabular}

\vspace{2pt}
\begin{tabular}{c}
(a) Pre-training setting.
\end{tabular}
\vspace{4pt}

\tablestyle{3pt}{1.02}
\begin{tabular}{y{80}|x{29}x{29}x{29}x{29}}
 & \multicolumn{3}{c}{ImageNet} & \multicolumn{1}{c}{Kinetics} \\
config & ViT-B & ViT-L & ViT-H & ViT-L \\
\shline
optimizer & \multicolumn{4}{c}{AdamW~\cite{adamw}} \\
optimizer momentum & \multicolumn{4}{c}{$\beta_1, \beta_2{=}0.9, 0.999$} \\
weight decay & \multicolumn{4}{c}{0.05} \\
learning rate schedule & \multicolumn{4}{c}{cosine decay~\cite{sgdr}} \\
warmup epochs~\cite{1hour} & \multicolumn{4}{c}{5} \\
augmentation & \multicolumn{4}{c}{\texttt{RandAug} (9, 0.5)~\cite{randaug}} \\
mixup~\cite{mixup} & \multicolumn{4}{c}{0.8} \\
cutmix~\cite{cutmix} & \multicolumn{4}{c}{1.0} \\
label smoothing~\cite{inception} & \multicolumn{4}{c}{0.1} \\
end lr & \multicolumn{4}{c}{1.0e-6} \\
batch size & \multicolumn{3}{c}{1024} & 128 \\
base lr & 5.0e-4 & 1.0e-3 & 1.0e-3 & 3.2e-3 \\
layer decay~\cite{electra} & 0.65 & 0.7 & 0.75 & 0.75 \\
base lr (w/ CLIP) & \multicolumn{3}{c}{2.0e-4} & 8.0e-4 \\
layer decay (w/ CLIP) & 0.65 & 0.75 & 0.8 & 0.8 \\
training epochs & 100 & 50 & 50 & 50 \\
drop path~\cite{droppath} & 0.1 & 0.1 & 0.3 & 0.2 \\
drop out~\cite{dropout} & \multicolumn{3}{c}{-} & 0.5 \\
repeated aug.~\cite{repeatedaug} & \multicolumn{3}{c}{-} & 2 \\
\end{tabular}

\vspace{2pt}
\begin{tabular}{c}
(b) Fine-tuning setting.
\end{tabular}
\vspace{4pt}

\caption{\small \textbf{Configurations on IN-1K and Kinetics-400.} In terms of learning rate (\textit{lr}), we use the linear scaling rule introduced in ~\cite{1hour}: \textit{lr} = \textit{base\_lr}$\times$\textit{batch\_size} / 256. When using repeated augmentation, the number of epochs and batch size count the original samples without repeating.}
\label{tab:supp-impl-in}
\vspace{-5pt}
\end{table}

During fine-tuning, we extract features from the encoder. We use global average pooling to gather the patch features, followed by a layer normalization and a linear classification head. Both layer normalization and the linear head are randomly initialized. Particularly, the linear head is initialized with a very small standard deviation $2^{-5}$, which enhances stability of fine-tuning.

\paragraph{Training recipes.} The default settings for pre-training and fine-tuning are in \cref{tab:supp-impl-in}. We use a different base learning rate and layer decay when fine-tuning CLIP-aided models.

\begin{table}[!t]
\tablestyle{3pt}{1.02}
\begin{tabular}{y{80}|x{65}x{65}}
config & from-scratch & fine-tuning \\
\shline
optimizer & \multicolumn{2}{c}{AdamW~\cite{adamw}} \\
optimizer momentum & \multicolumn{2}{c}{$\beta_1, \beta_2{=}0.9, 0.999$} \\
weight decay & \multicolumn{2}{c}{0.02} \\
learning rate schedule & \multicolumn{2}{c}{cosine decay~\cite{sgdr}} \\
warmup epochs~\cite{1hour} & \multicolumn{2}{c}{10} \\
augmentation & \multicolumn{2}{c}{\texttt{RandAug}(9, 0.5)} \\
mixup~\cite{mixup} & \multicolumn{2}{c}{0.8} \\
cutmix~\cite{cutmix} & \multicolumn{2}{c}{1.0} \\
label smoothing~\cite{inception} & \multicolumn{2}{c}{0.1} \\
batch size & \multicolumn{2}{c}{512} \\
epochs & \multicolumn{2}{c}{200} \\
base lr & \multicolumn{2}{c}{1.0e-3} \\
layer decay~\cite{electra} & - & 0.8 \\
drop path~\cite{droppath} & 0.1 & 0.2 \\ 
\end{tabular}
\vspace{2pt}
\caption{\small \textbf{Configurations of fine-tuning ADM~\cite{adm} on IN-1K.} \\In terms of learning rate (\textit{lr}), we use the linear scaling rule introduced in ~\cite{1hour}: \textit{lr} = \textit{base\_lr}$\times$\textit{batch\_size} / 256. For fine-tuning, we use ADM's unconditional $256^2$ model trained on IN-1K.}
\label{tab:supp-adm}
\vspace{-5pt}
\end{table}

\subsection{Kinetics Experiments}

\paragraph{Architecture.} Given a video clip, we first divide it into non-overlapping patches in spacetime. Positional embeddings are added to the embedded patches. The spacetime patch size is $2$\,\x\,$16$\,\x\,$16$ for ViT-L/16 and $2$\,\x\,$14$\,\x\,$14$ for ViT-L/14. The target of our \ourmodel is a single time slice of the patch ($16$\,\x\,$16$ or $14$\,\x\,$14$), and so are the corresponding noisy inputs to the decoder~\cite{videomae}. Similar to the image setting, the encoder and the decoder use two different linear projections to handle the clean and the noisy (masked) inputs, respectively. We use 90\% random masking sampling on the spacetime patches~\cite{videomae}. 

We extract features from the encoder outputs for fine-tuning. We use global average pooling to gather the patch features, followed by a layer normalization and a linear head. The linear head is initialized with a very small standard deviation $2^{-5}$, the same as the image setting. To further enhance the results, we fine-tune the $16$\,\x\,$224^2$ Kinetics-400 model to a longer duration $32$ and a larger resolution $280^2$ for a short schedule of 30 epochs without repeated augmentation.

\paragraph{Training recipes.} The default settings for pre-training and fine-tuning are in \cref{tab:supp-impl-in}. Note that many hyper-parameters are shared by the image and the video models, showing that \ourmodel is general across different domains. We search for the best base learning rate and layer decay when fine-tuning CLIP-aided models.

\subsection{Fine-Tuning ADM}
We fine-tune the pre-trained ADM~\cite{adm} model to evaluate the recognition ability of this well-designed diffusion model. Specifically, we take its IN-1K unconditional 256$^2$ version and fine-tune the model at resolution 224$^2$ on IN-1K classification for a fair comparison to other methods.

The ADM model uses a U-Net~\cite{unet} architecture for dense prediction. It consists of ResNet~\cite{resnet} blocks and self-attention layers~\cite{transformer}. We fine-tune the input blocks and the middle block, which are followed by a global average pooling, a layer normalization, and a linear classification head that projects the global averaged feature to classification logits. The layer normalization and the linear head are randomly initialized. Regarding the timestep input specifying the noise level for diffusion generation, we simply fix the timestep to 999 for classification fine-tuning, while other numbers that are inside the range of the noise schedule, \ie, from 0 to 999, give similar training curves and results. We also train the same model from scratch as the baseline to show the effectiveness of diffusion generative pre-training.

\paragraph{Training recipes.} We include the training recipes of fine-turning and from-scratch training of ADM in \cref{tab:supp-adm}. We carefully tune the optimization hyper-parameters of both the fine-tuning and the from-scratch training. The recipes are based on sophisticated modern training techniques~\cite{deit, convnext}, and we tune base learning rate, layer-wise decay~\cite{electra}, and drop path rate for each case.

\section{Additional Qualitative Results}
\label{sec:more-viz}

\begin{figure*}[!ht]
\centering
\tablestyle{0.5pt}{0.2}
\begin{tabular}{cc}
\includegraphics[width=0.5\textwidth]{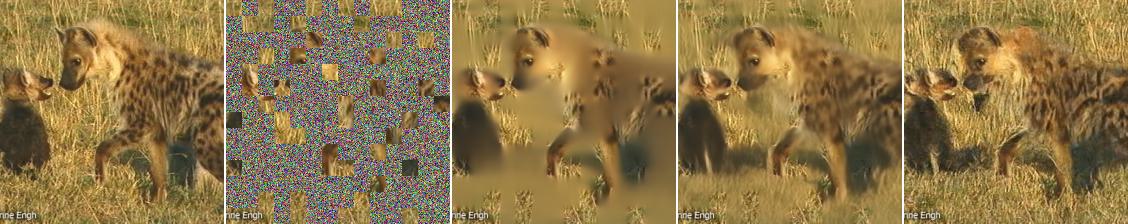} &
\includegraphics[width=0.5\textwidth]{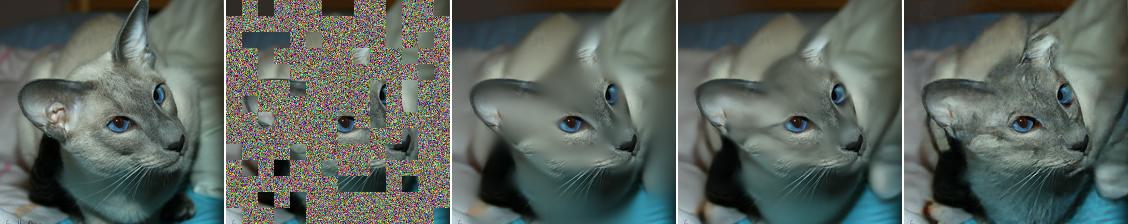} \\
\includegraphics[width=0.5\textwidth]{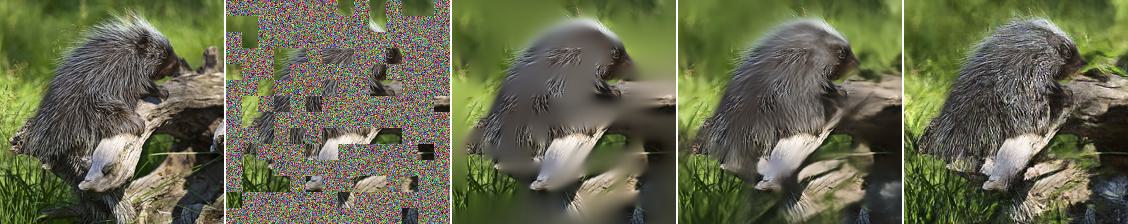} &
\includegraphics[width=0.5\textwidth]{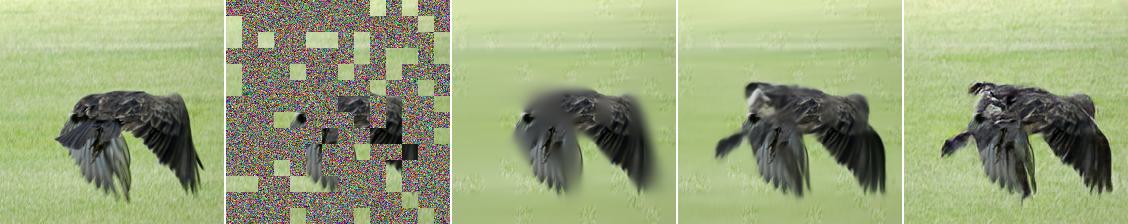} \\
\includegraphics[width=0.5\textwidth]{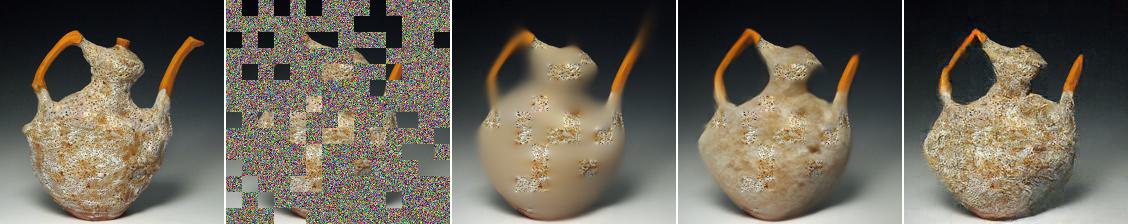} &
\includegraphics[width=0.5\textwidth]{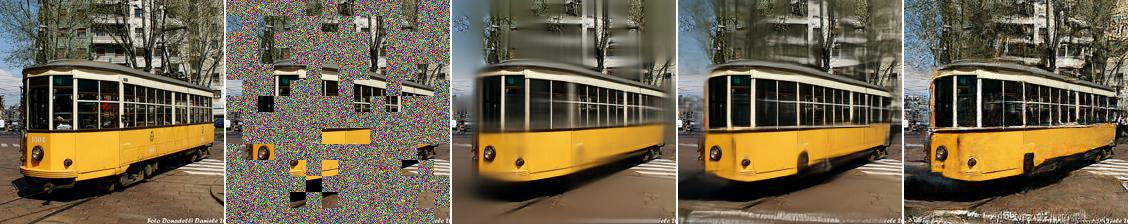} \\
\includegraphics[width=0.5\textwidth]{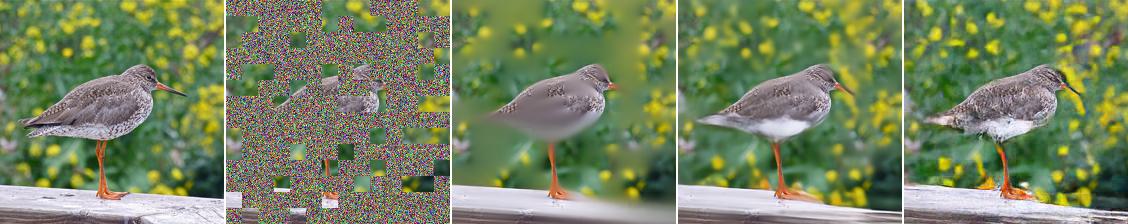} &
\includegraphics[width=0.5\textwidth]{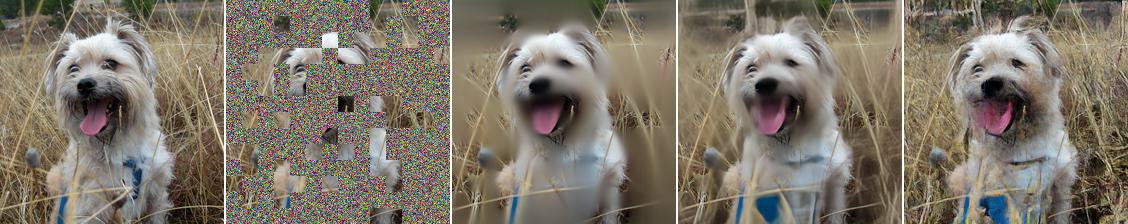} \\
\includegraphics[width=0.5\textwidth]{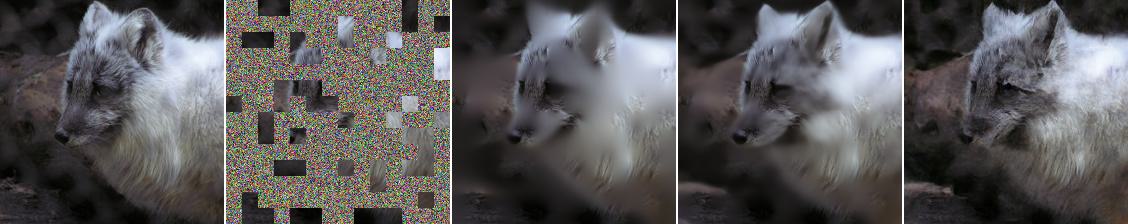} &
\includegraphics[width=0.5\textwidth]{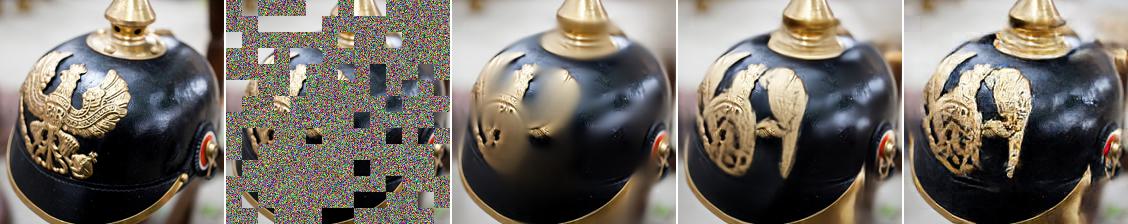} \\
\includegraphics[width=0.5\textwidth]{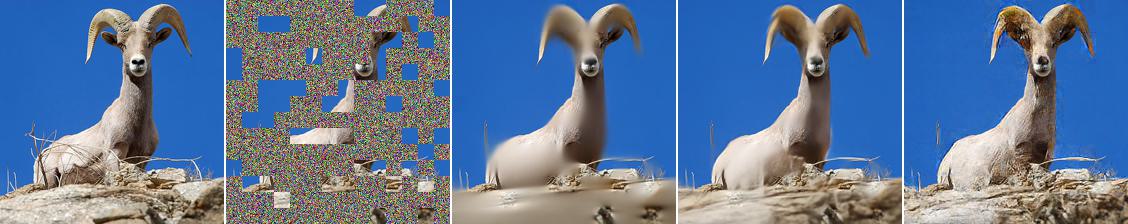} &
\includegraphics[width=0.5\textwidth]{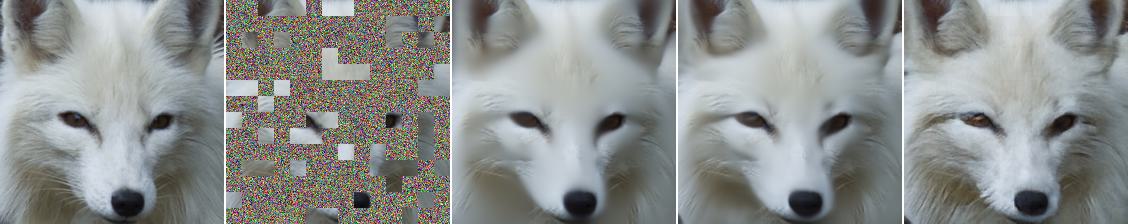} \\
\includegraphics[width=0.5\textwidth]{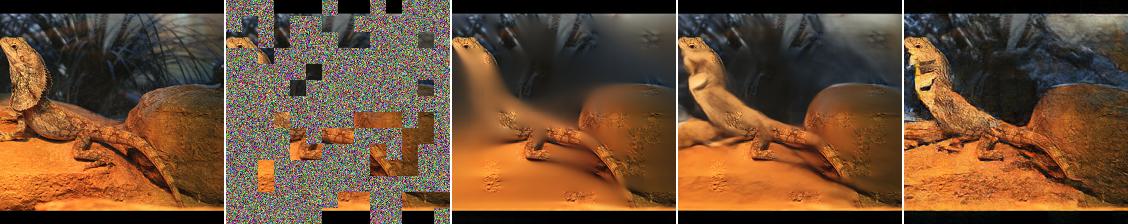} &
\includegraphics[width=0.5\textwidth]{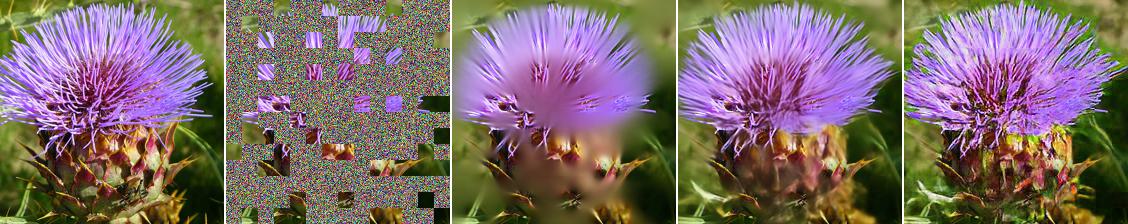} \\
\includegraphics[width=0.5\textwidth]{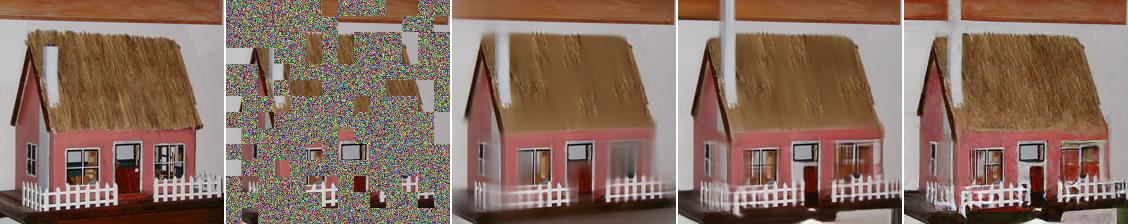} &
\includegraphics[width=0.5\textwidth]{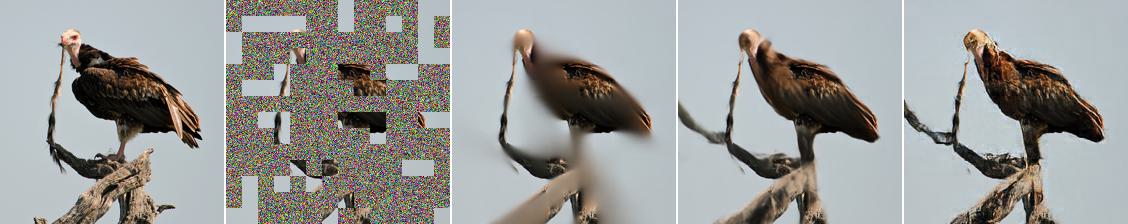} \\
\includegraphics[width=0.5\textwidth]{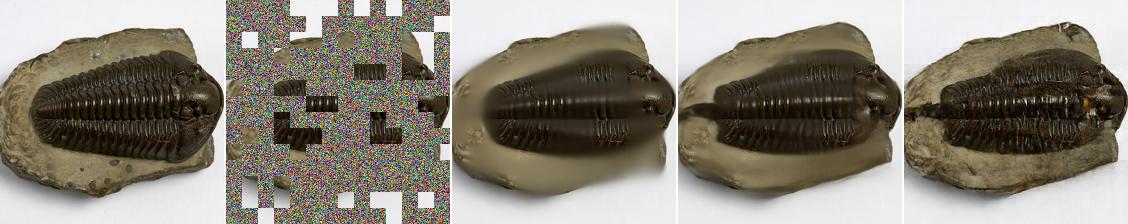} &
\includegraphics[width=0.5\textwidth]{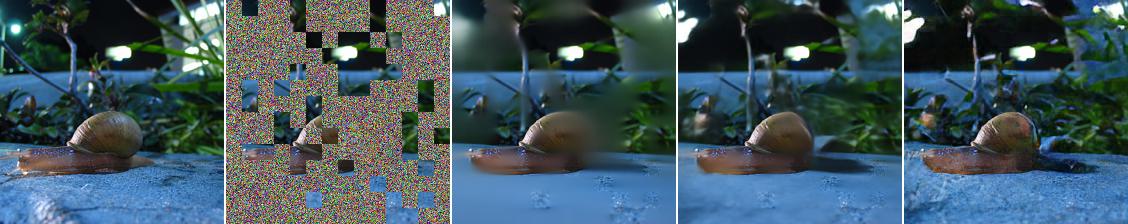} \\
\includegraphics[width=0.5\textwidth]{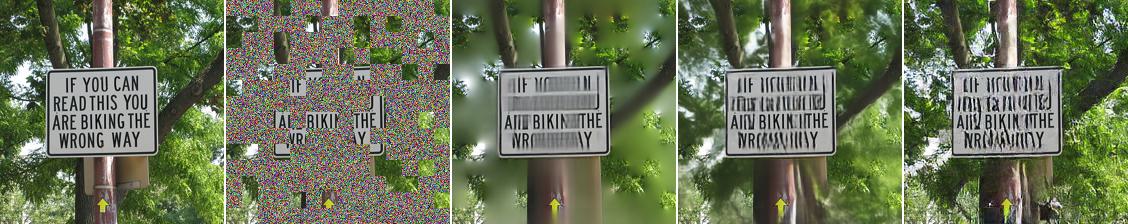} &
\includegraphics[width=0.5\textwidth]{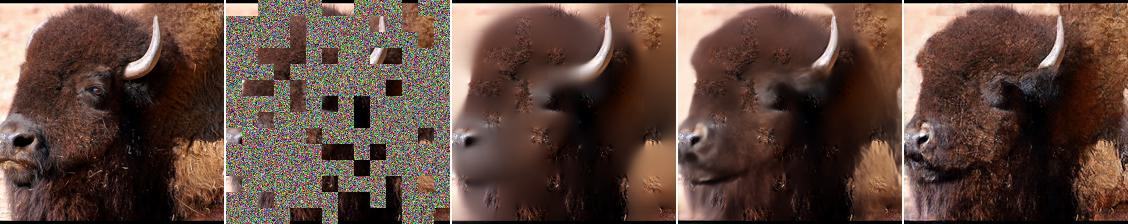} \\ 
\includegraphics[width=0.5\textwidth]{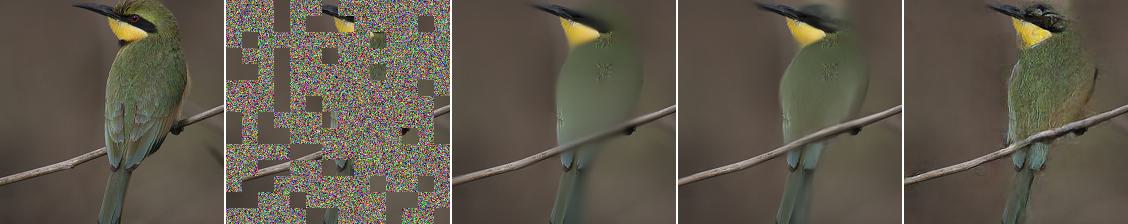} & 
\includegraphics[width=0.5\textwidth]{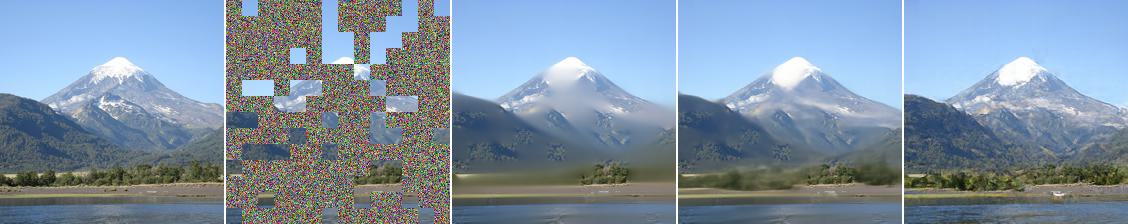} \\
\includegraphics[width=0.5\textwidth]{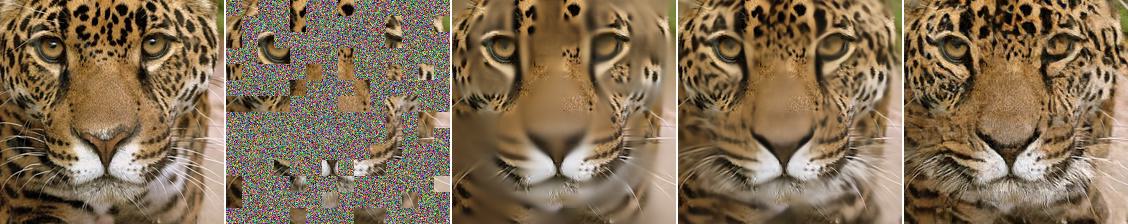} & 
\includegraphics[width=0.5\textwidth]{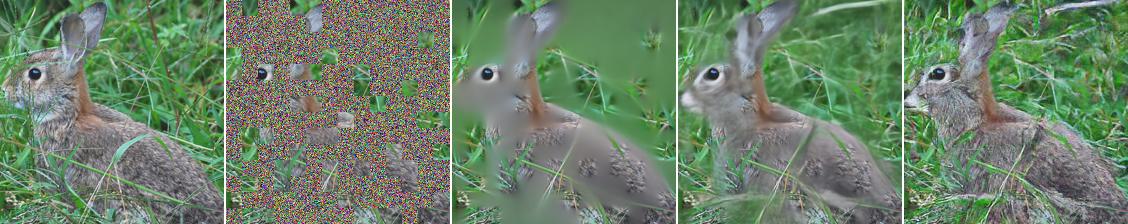} \\
\end{tabular}

\begin{tabular}{x{49}x{49}x{49}x{49}x{49}x{49}x{49}x{49}x{49}x{49}}
\footnotesize{ground-truth} & \footnotesize{masked} & \footnotesize{$t=1000$} & \footnotesize{$t=500$} & \footnotesize{\textbf{$t=0$}} & \footnotesize{ground-truth} & \footnotesize{masked} & \footnotesize{$t=1000$} & \footnotesize{$t=500$} & \footnotesize{\textbf{$t=0$}}
\end{tabular}
\caption{\textbf{Visualizations of \ourmodel generation with 75\% random masking.} The images are from IN-1K validation set with size 224$^2$. We show the reverse diffusion at $t$\,$=$\,$1000$, $500$, and $0$. $t$\,$=$\,$0$ is the final output. The model is ViT-L. Best viewed in color with zoom.}
\label{fig:supp-in-random-1}
\end{figure*}

\begin{figure*}[!ht]
\centering
\tablestyle{0.5pt}{0.2}
\begin{tabular}{cc}
\includegraphics[width=0.5\textwidth]{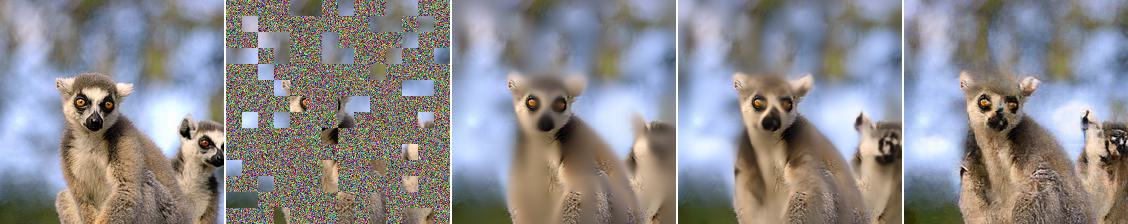} &
\includegraphics[width=0.5\textwidth]{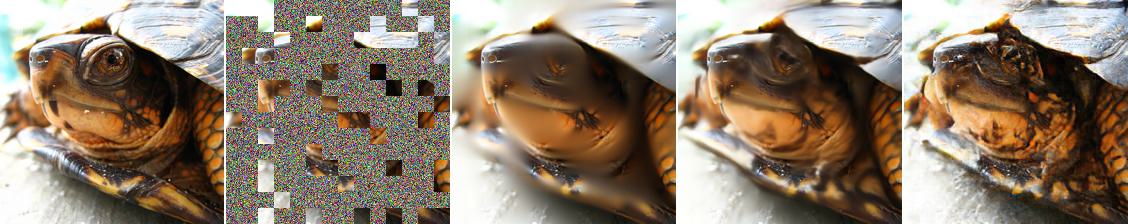} \\
\includegraphics[width=0.5\textwidth]{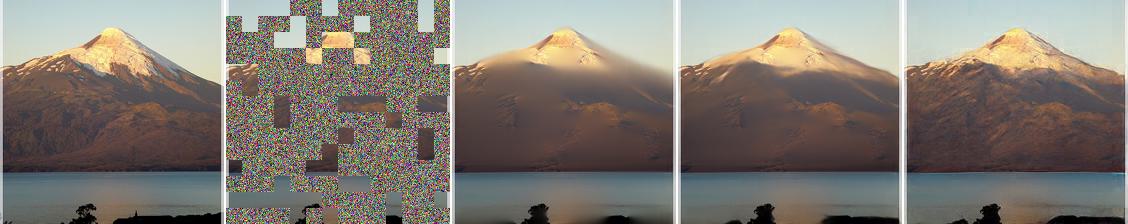} &
\includegraphics[width=0.5\textwidth]{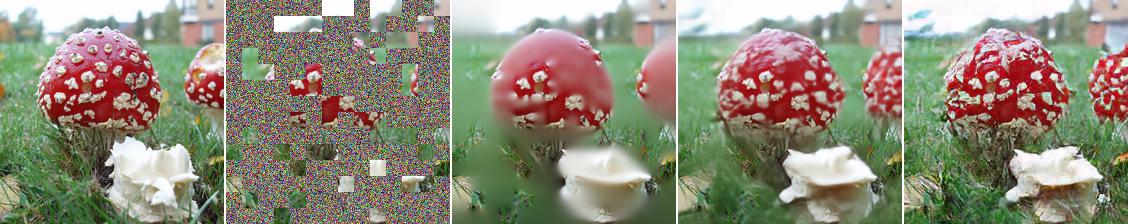} \\
\includegraphics[width=0.5\textwidth]{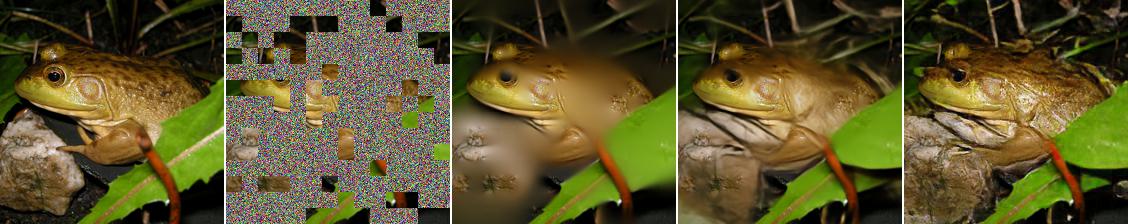} &
\includegraphics[width=0.5\textwidth]{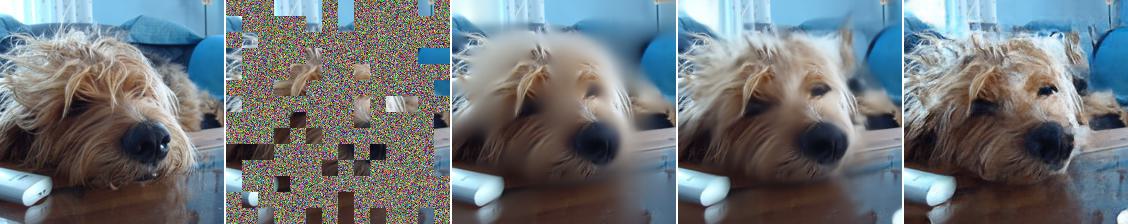} \\
\includegraphics[width=0.5\textwidth]{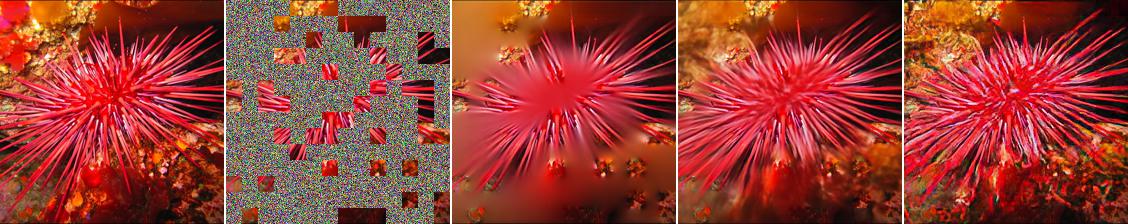} &
\includegraphics[width=0.5\textwidth]{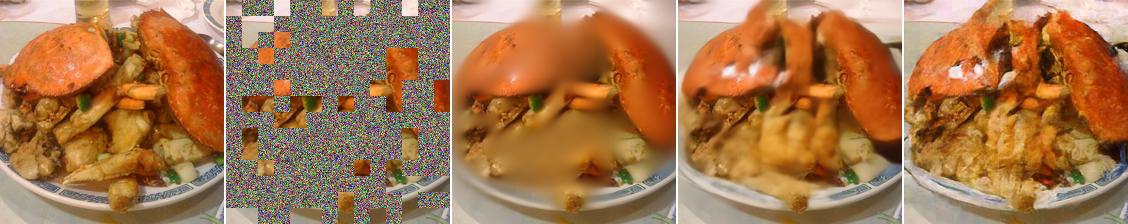} \\
\includegraphics[width=0.5\textwidth]{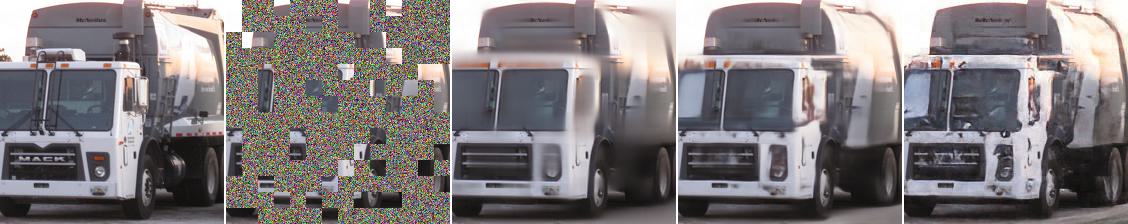} &
\includegraphics[width=0.5\textwidth]{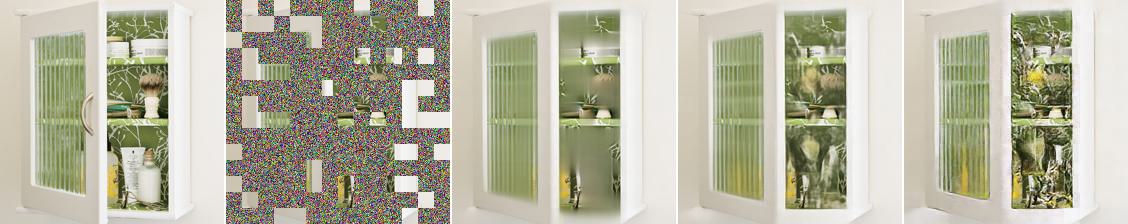} \\
\includegraphics[width=0.5\textwidth]{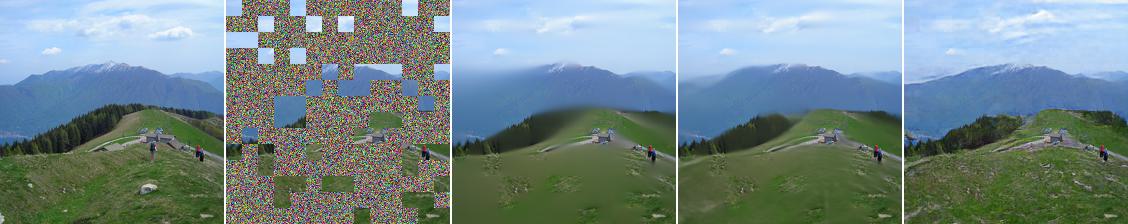} &
\includegraphics[width=0.5\textwidth]{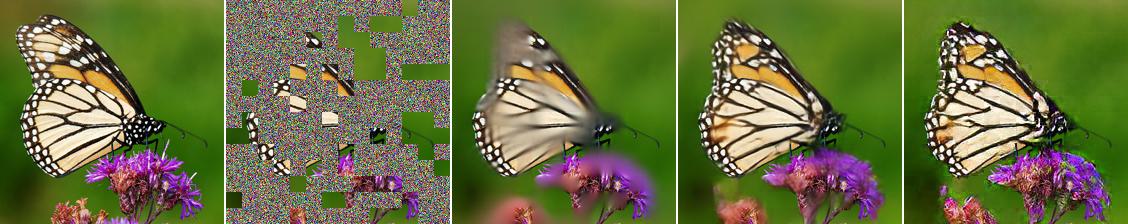} \\
\includegraphics[width=0.5\textwidth]{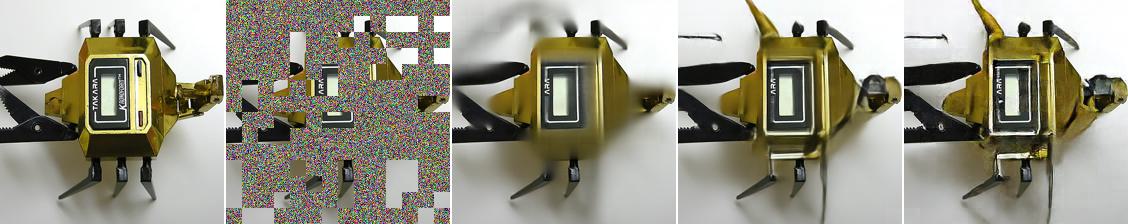} &
\includegraphics[width=0.5\textwidth]{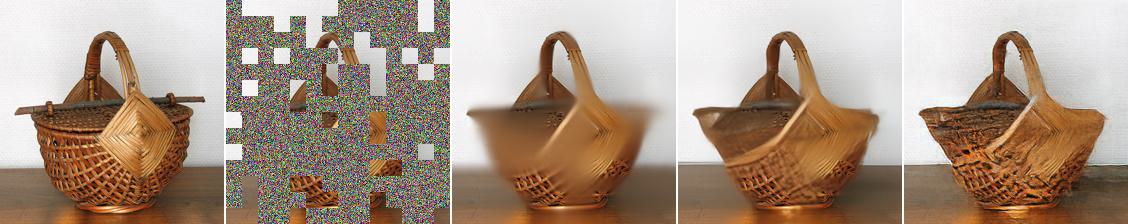} \\
\includegraphics[width=0.5\textwidth]{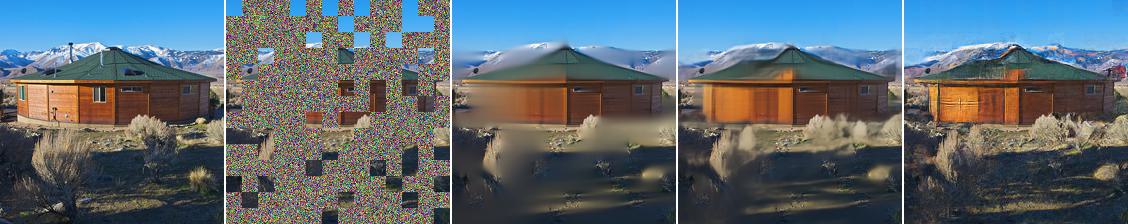} &
\includegraphics[width=0.5\textwidth]{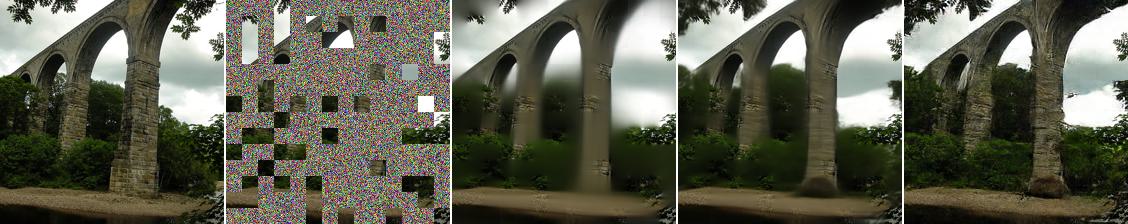} \\
\includegraphics[width=0.5\textwidth]{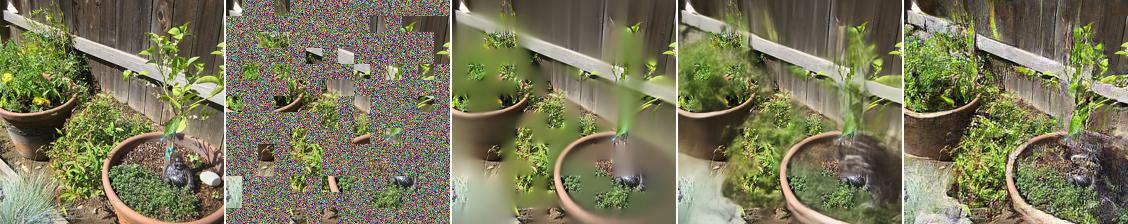} &
\includegraphics[width=0.5\textwidth]{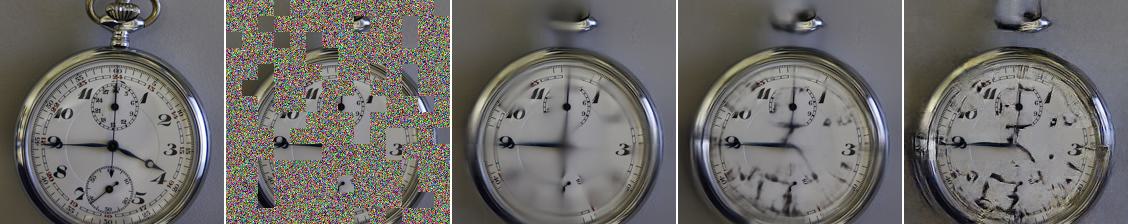} \\
\includegraphics[width=0.5\textwidth]{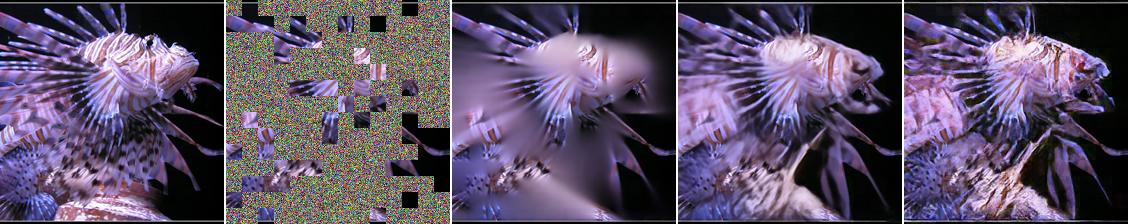} &
\includegraphics[width=0.5\textwidth]{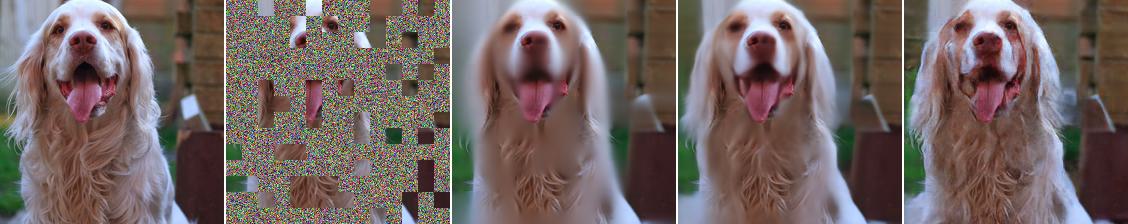} \\
\includegraphics[width=0.5\textwidth]{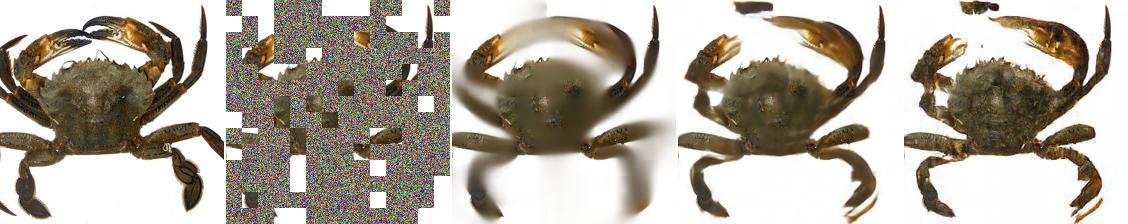} &
\includegraphics[width=0.5\textwidth]{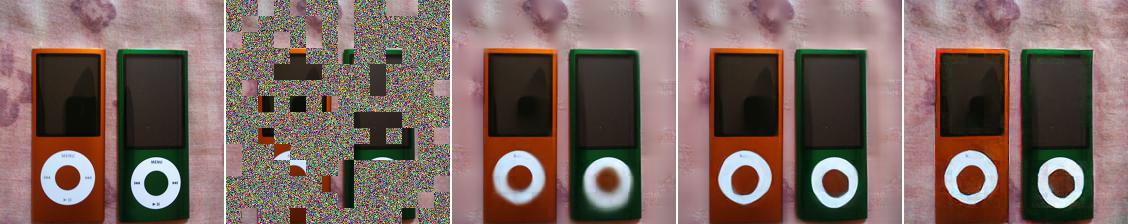} \\
\includegraphics[width=0.5\textwidth]{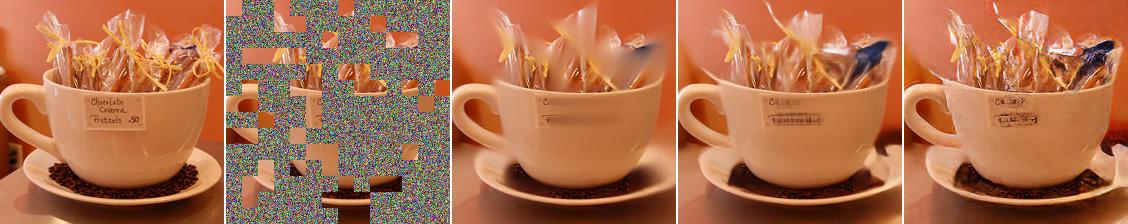} &
\includegraphics[width=0.5\textwidth]{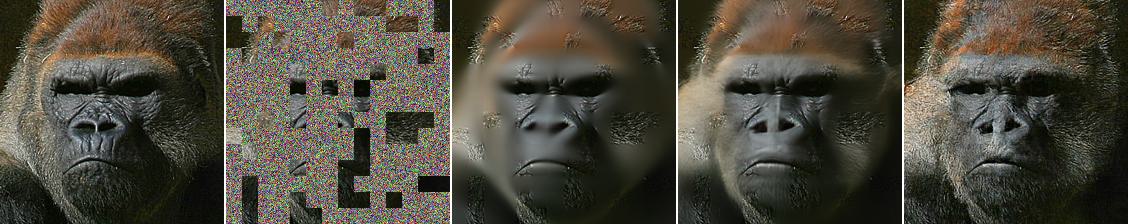} \\
\end{tabular}

\begin{tabular}{x{49}x{49}x{49}x{49}x{49}x{49}x{49}x{49}x{49}x{49}}
\footnotesize{ground-truth} & \footnotesize{masked} & \footnotesize{$t=1000$} & \footnotesize{$t=500$} & \footnotesize{\textbf{$t=0$}} & \footnotesize{ground-truth} & \footnotesize{masked} & \footnotesize{$t=1000$} & \footnotesize{$t=500$} & \footnotesize{\textbf{$t=0$}}
\end{tabular}

\caption{\textbf{Visualizations of \ourmodel generation with 75\% random masking.} The images are from IN-1K validation set with size 224$^2$. We show the reverse diffusion at $t$\,$=$\,$1000$, $500$, and $0$. $t$\,$=$\,$0$ is the final output. The model is ViT-L. Best viewed in color with zoom.}
\label{fig:supp-in-random-2}
\end{figure*}

\begin{figure*}[!ht]
\centering
\tablestyle{0.5pt}{0.0}
\begin{tabular}{cc}
\includegraphics[width=0.45\textwidth]{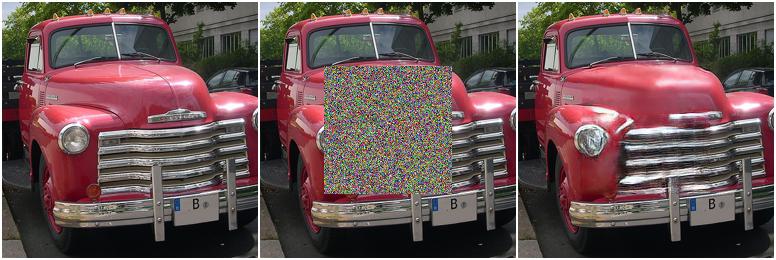} &
\includegraphics[width=0.45\textwidth]{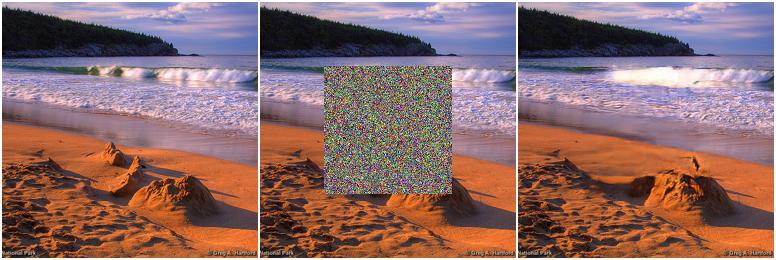} \\
\includegraphics[width=0.45\textwidth]{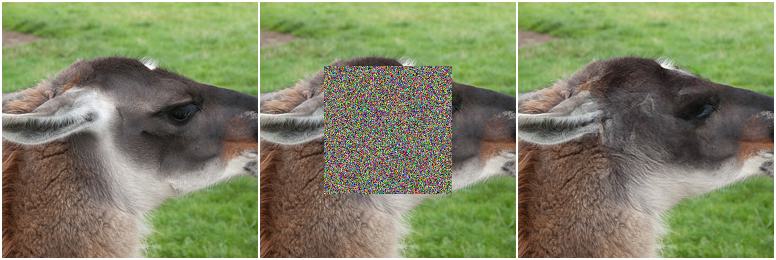} & 
\includegraphics[width=0.45\textwidth]{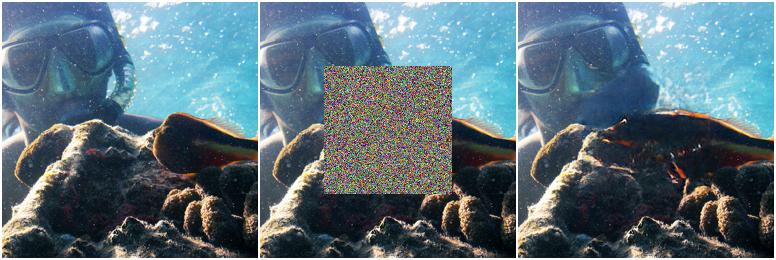} \\
\includegraphics[width=0.45\textwidth]{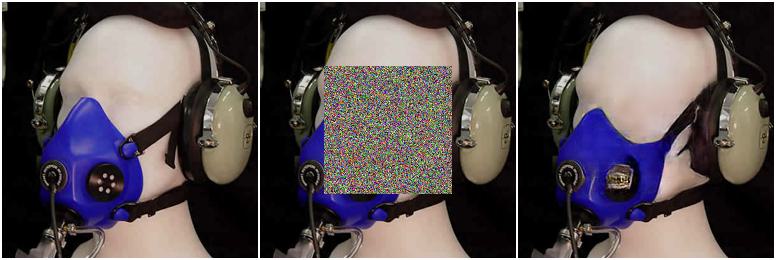} &
\includegraphics[width=0.45\textwidth]{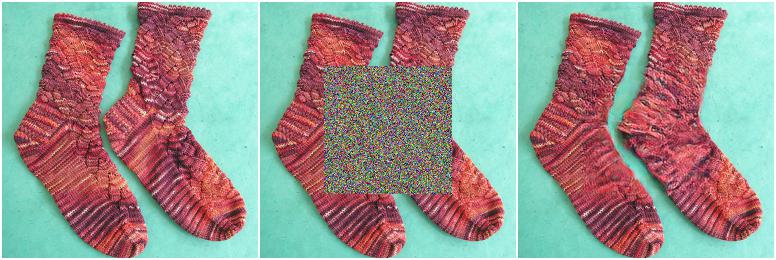} \\
\includegraphics[width=0.45\textwidth]{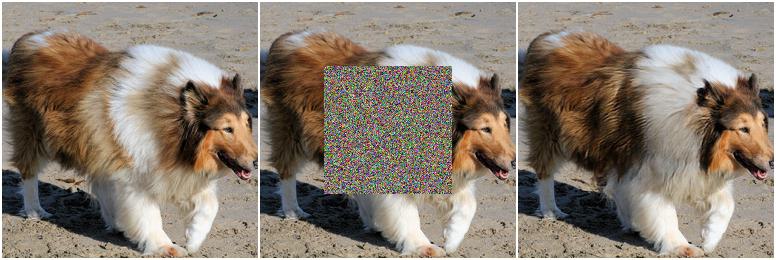} &
\includegraphics[width=0.45\textwidth]{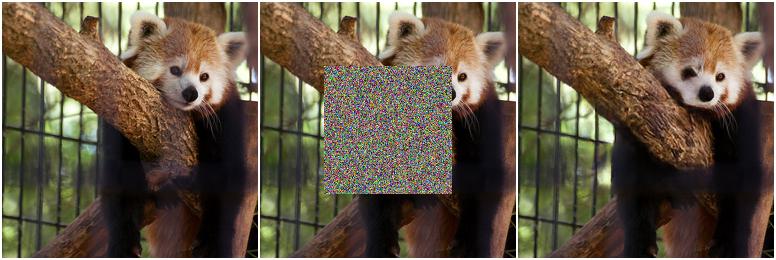} \\
\includegraphics[width=0.45\textwidth]{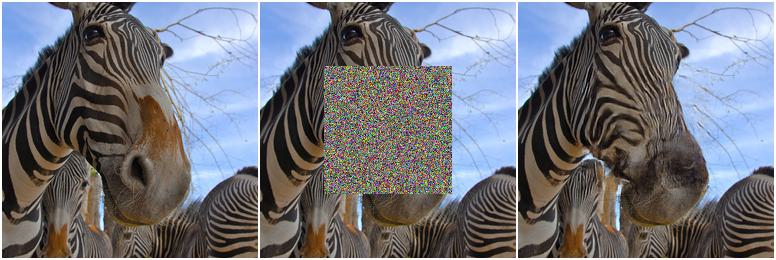} &
\includegraphics[width=0.45\textwidth]{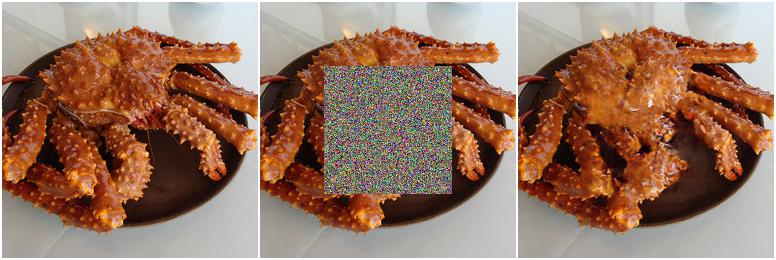} \\
\includegraphics[width=0.45\textwidth]{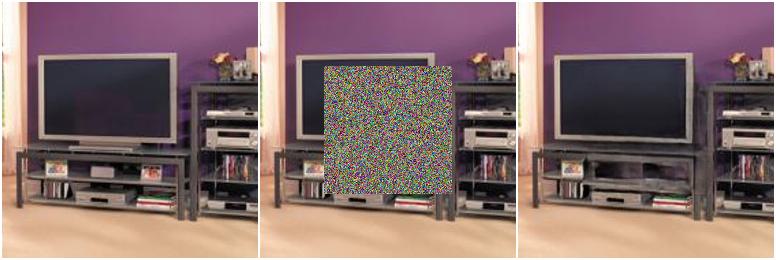} &
\includegraphics[width=0.45\textwidth]{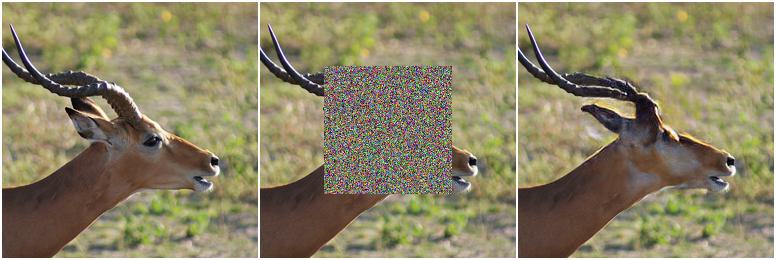} \\
\includegraphics[width=0.45\textwidth]{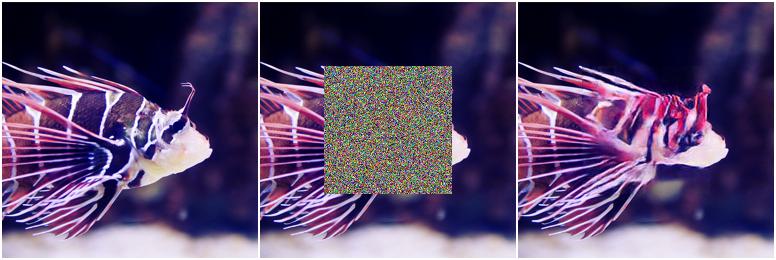} &
\includegraphics[width=0.45\textwidth]{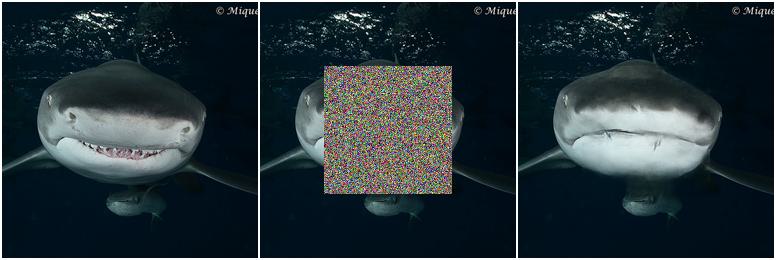} \\
\includegraphics[width=0.45\textwidth]{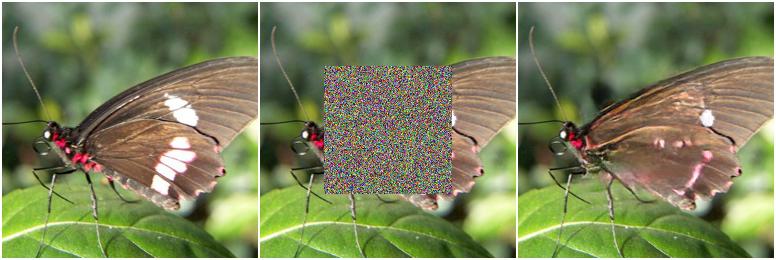} &
\includegraphics[width=0.45\textwidth]{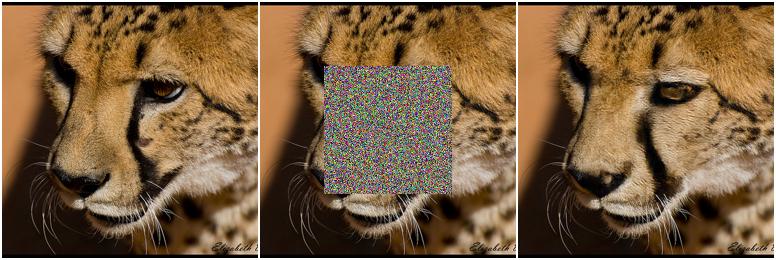} \\
\end{tabular}

\begin{tabular}{x{78}x{78}x{78}x{78}x{78}x{78}}
\footnotesize{ground-truth} & \footnotesize{masked} & \footnotesize{generated} & \footnotesize{ground-truth} & \footnotesize{masked} & \footnotesize{generated}
\end{tabular}
\caption{\textbf{Visualizations of \ourmodel generation with center masking.} The images are from IN-1K validation set. The input images are of size 256$^2$, with the center 128$^2$ block masked. The model is ViT-L. Best viewed in color with zoom.}
\label{fig:supp-in-center}
\end{figure*}

\begin{figure*}[!ht]
\centering
\tablestyle{0.5pt}{0.0}
\begin{tabular}{cc}
\includegraphics[width=0.45\textwidth]{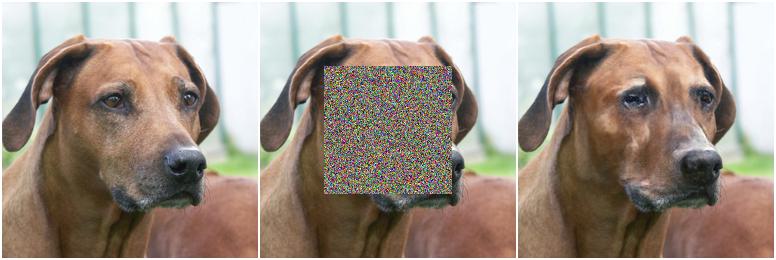} &
\includegraphics[width=0.45\textwidth]{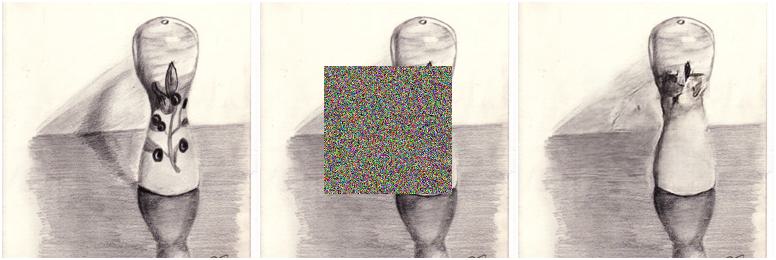} \\
\includegraphics[width=0.45\textwidth]{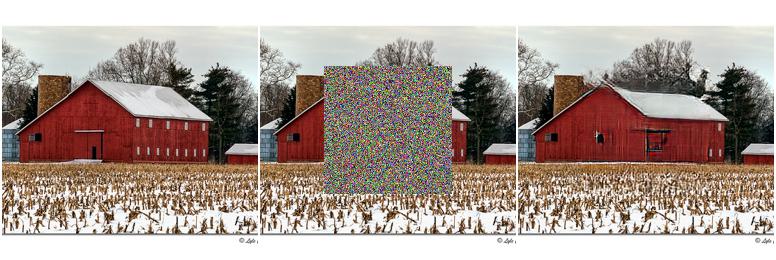} & 
\includegraphics[width=0.45\textwidth]{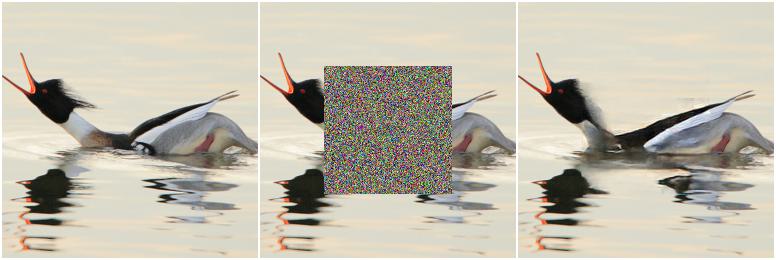} \\
\includegraphics[width=0.45\textwidth]{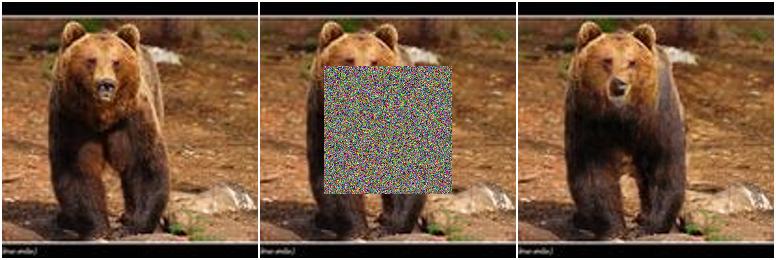} &
\includegraphics[width=0.45\textwidth]{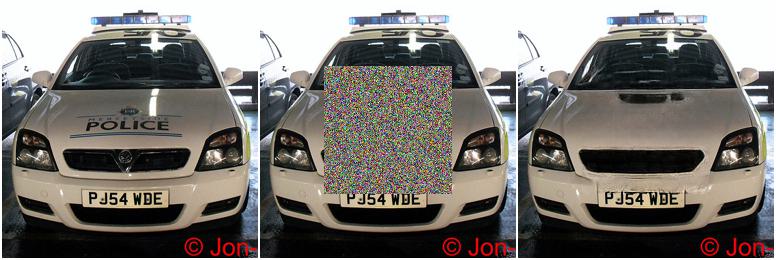} \\
\includegraphics[width=0.45\textwidth]{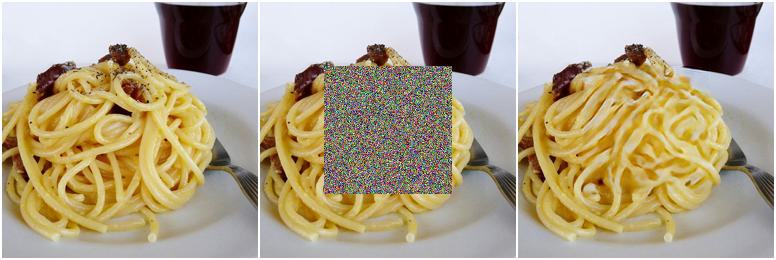} &
\includegraphics[width=0.45\textwidth]{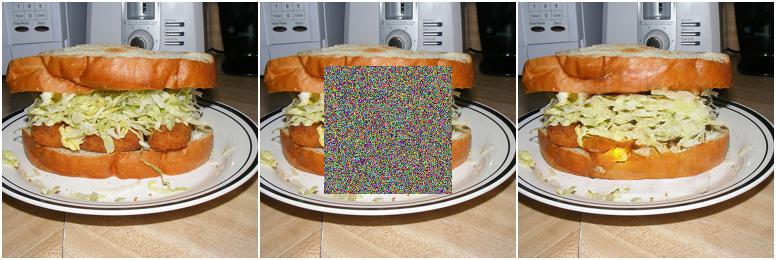} \\
\includegraphics[width=0.45\textwidth]{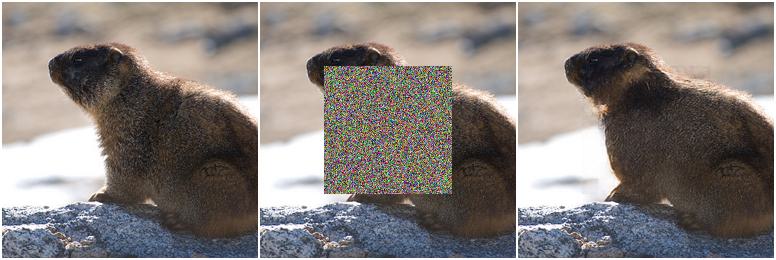} &
\includegraphics[width=0.45\textwidth]{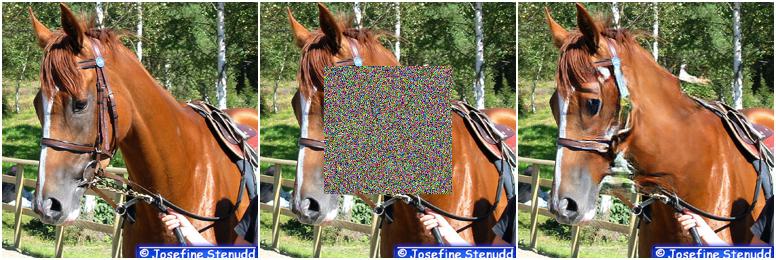} \\
\includegraphics[width=0.45\textwidth]{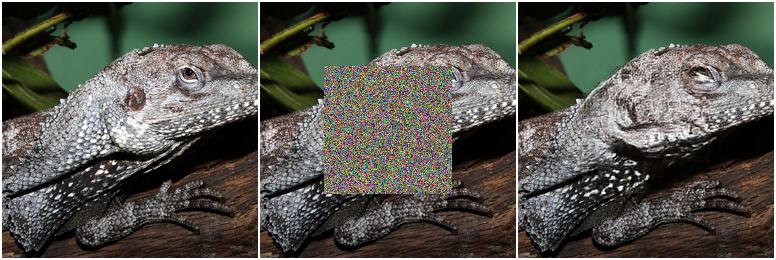} &
\includegraphics[width=0.45\textwidth]{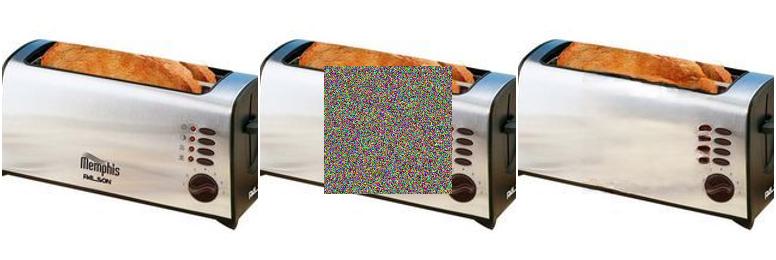} \\
\includegraphics[width=0.45\textwidth]{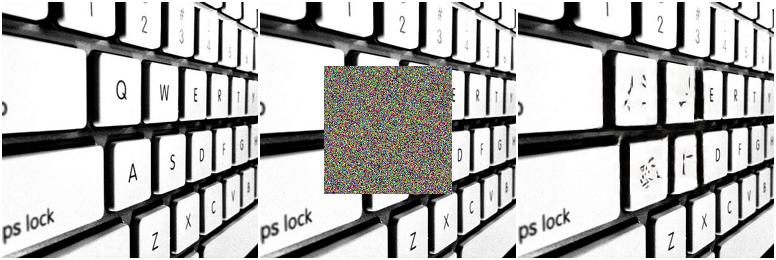} &
\includegraphics[width=0.45\textwidth]{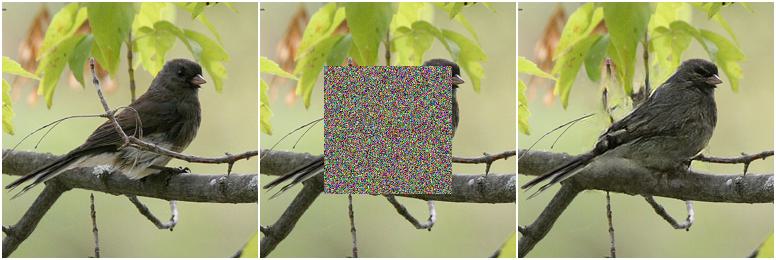} \\
\includegraphics[width=0.45\textwidth]{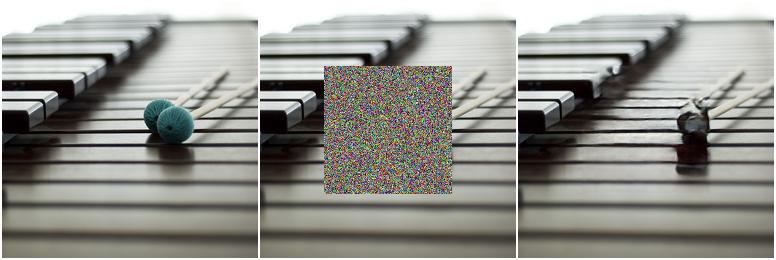} &
\includegraphics[width=0.45\textwidth]{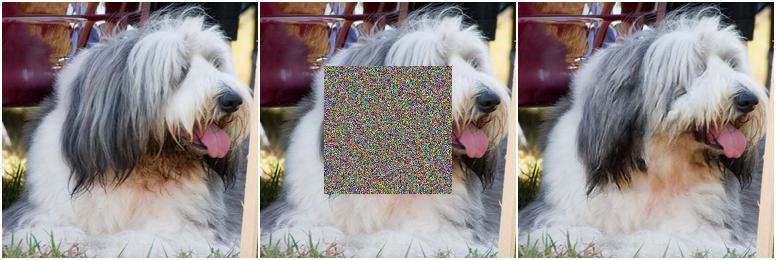} \\
\end{tabular}

\begin{tabular}{x{78}x{78}x{78}x{78}x{78}x{78}}
\footnotesize{ground-truth} & \footnotesize{masked} & \footnotesize{generated} & \footnotesize{ground-truth} & \footnotesize{masked} & \footnotesize{generated}
\end{tabular}
\caption{\textbf{Visualizations of \ourmodel generation with center masking.} The images are from IN-1K validation set. The input images are of size 256$^2$, with the center 128$^2$ block masked. The model is ViT-L. Best viewed in color with zoom.}
\label{fig:supp-in-center-2}
\end{figure*}

We provide more qualitative results of image generation using ImageNet-1K validation images. \cref{fig:supp-in-random-1,fig:supp-in-random-2} are samples with 75\% random masking. \cref{fig:supp-in-center,fig:supp-in-center-2} are samples with the center block masking.

\begin{figure*}[!ht]
\centering
\tablestyle{0.5pt}{0.0}
\begin{tabular}{c}
\includegraphics[width=0.81\textwidth]{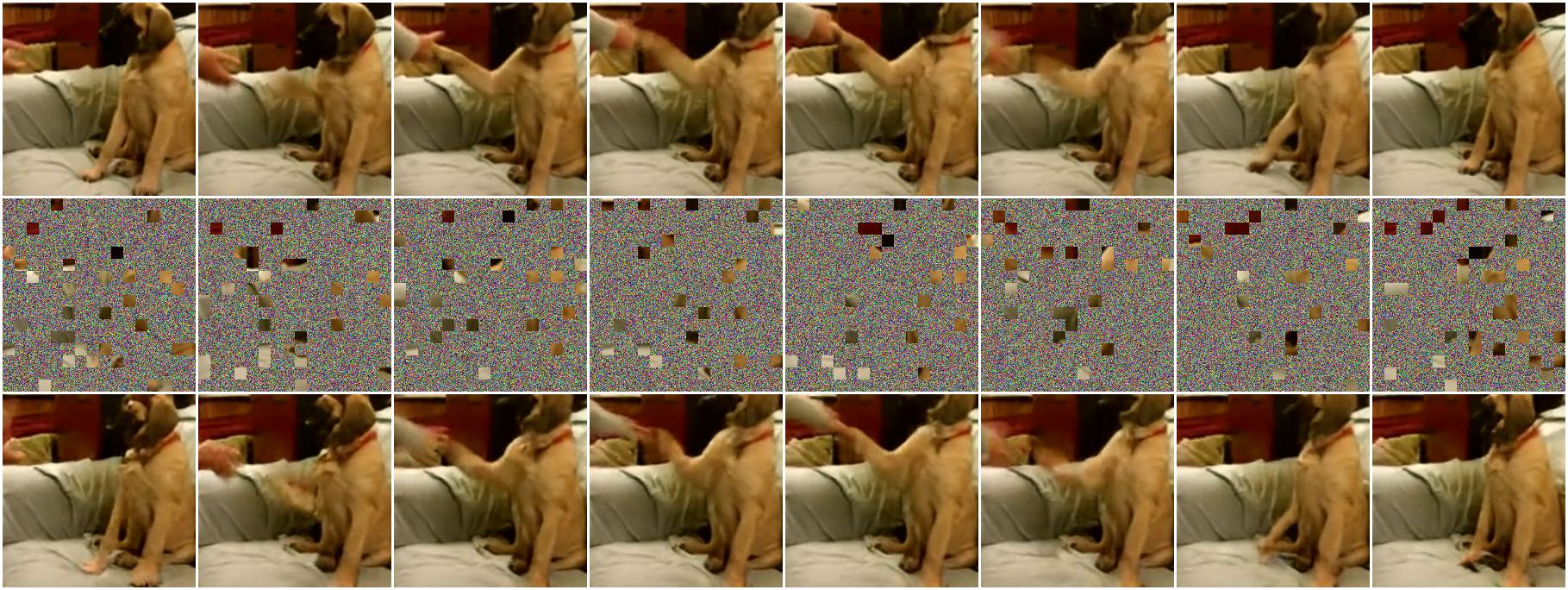} \\
\includegraphics[width=0.81\textwidth]{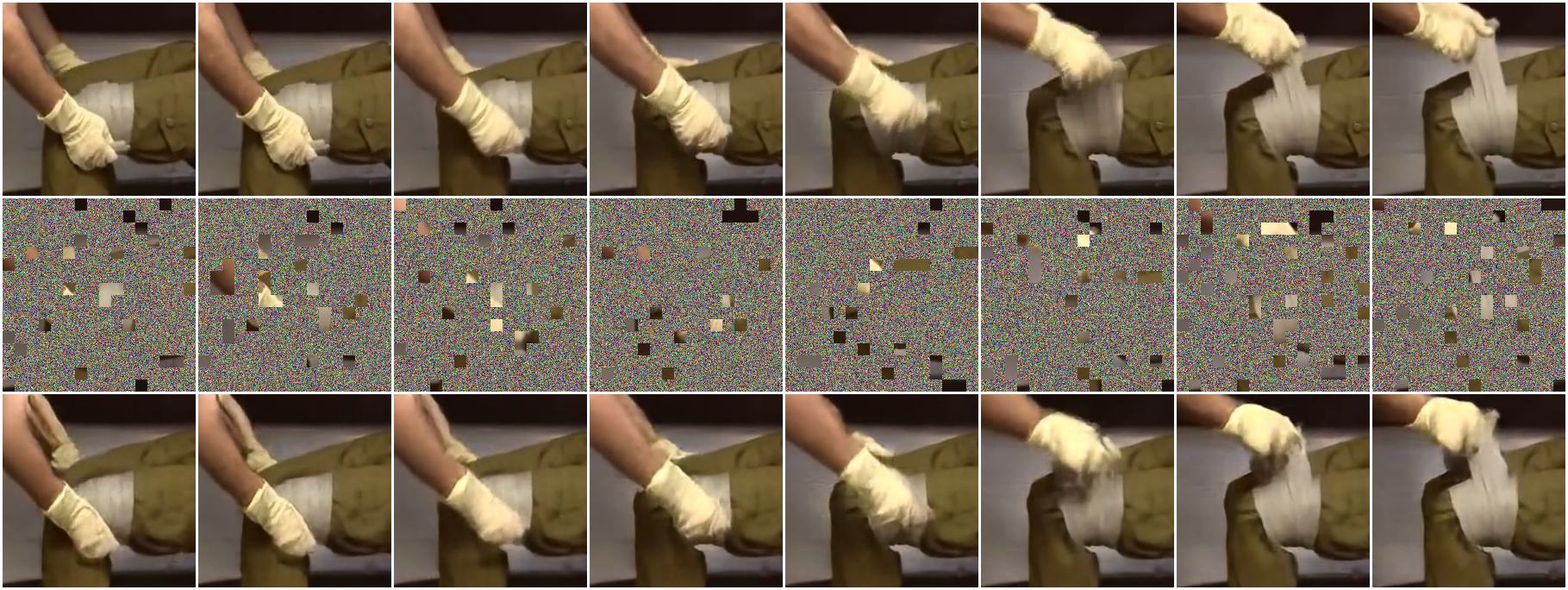} \\
\includegraphics[width=0.81\textwidth]{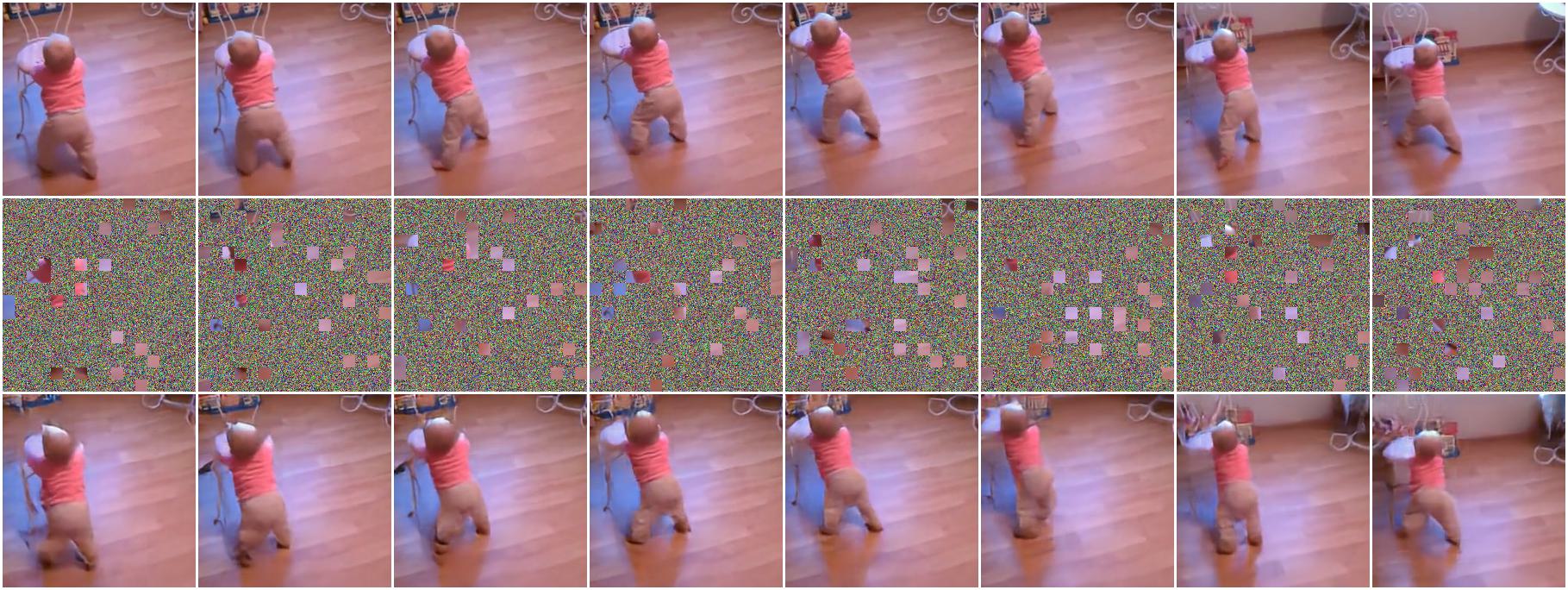} \\
\includegraphics[width=0.81\textwidth]{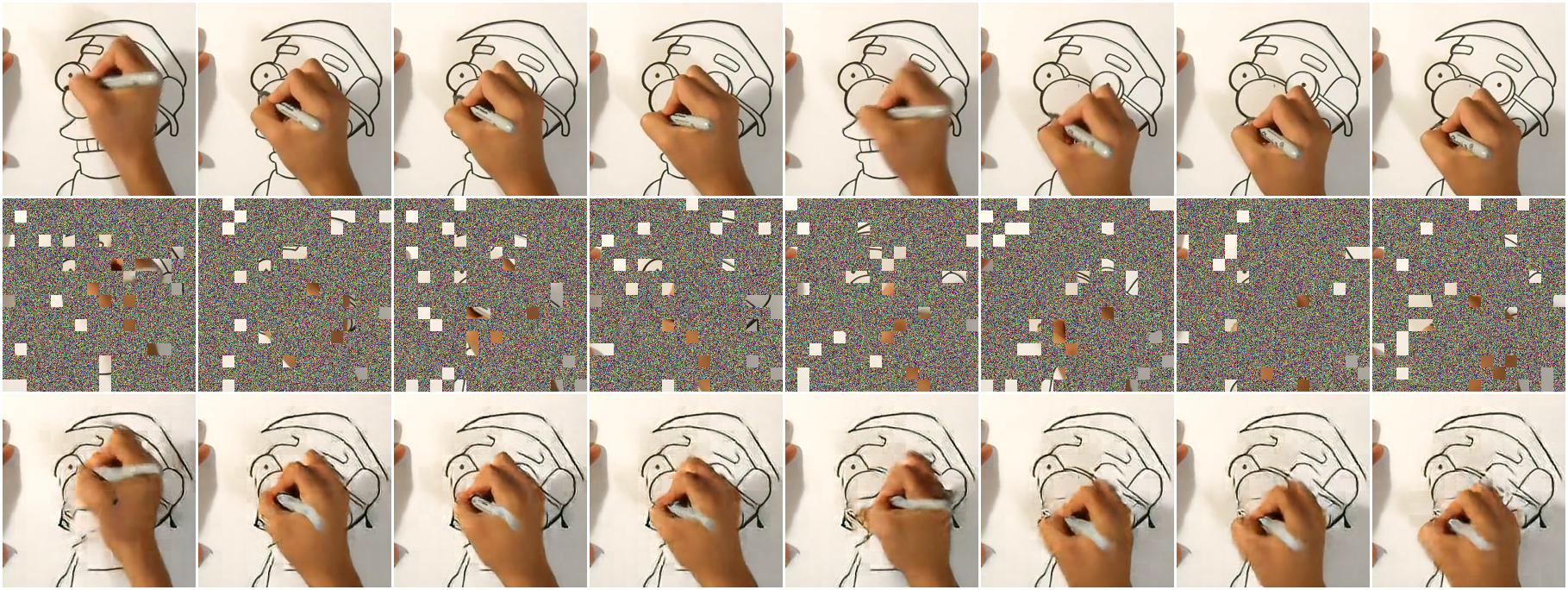} \\
\end{tabular}
\caption{\textbf{Visualizations of \ourmodel generation with video.} The videos are from Kinetics-400 validation set with random masking ratio 90\%. We show the original video (\textit{top}), masked video (\textit{middle}), and \ourmodel output (\textit{bottom}) for each sample. The model is ViT-L/14.}
\label{fig:supp-k400-1}
\end{figure*}

In \cref{fig:supp-k400-1}, we provide visualizations on \ourmodel for video generation on Kinetics-400 validation videos. For a $16$\,\x\,$224$\,\x\,$224$ video clip, we visualize the generated frames at stride two on the temporal dimension, which makes eight frames for each sample.

\paragraph{Acknowledgement.}
We thank Yuan-Ting Hu and Minkai Xu for valuable discussion.

\clearpage
{\small
\bibliographystyle{ieee_fullname}
\bibliography{egbib}
}

\end{document}